\documentclass[final,10pt,a4paper]{article}

\usepackage{t1enc}
\usepackage{graphicx}
\usepackage{tabularx}
\usepackage{color}

\usepackage{amsmath}
\usepackage{dsfont}

\usepackage[T1]{url}
\usepackage{achicago} \usepackage{abbrevs} \usepackage{lips}

\let\em\itswitch

\providecommand{\ie}{\textit{i.e.} }%
\providecommand{\eg}{\textit{e.g.} }%
\providecommand{\possessivecite}[1]{\citeANP{#1}'s~\citeyear{#1}}%

\newabbrev\mds{Multi-Document Summarization}[MDS]
\newabbrev\sds{Single-document Summarization}[SDS]
\newabbrev\rst{Rhetorical Structure Theory}[RST]
\newabbrev\sdrs{Synchronic and Diachronic Relations}[SDRs]
\newabbrev\nlg{Natural Language Generation}[NLG]
\newabbrev\tdt{Topic Detection and Tracking}[TDT]
\newabbrev\muc{Message Understanding Conferences (MUC)}[MUC]
\newabbrev\ml{Machine Learning}
\newabbrev\cst{Cross-do\-cu\-ment Structure Theory (CST)}[CST]

\begin{document}

\author{Stergos D. Afantenos\thanks{Corresponding author; email:
        \protect\url{stergos.afantenos@lif.univ-mrs.fr}} \footnotemark[2]
        \and Vangelis Karkaletsis\footnotemark[3]
        \and Panagiotis Stamatopoulos\footnotemark[4]
        \and Constantin Halatsis\footnotemark[4]}

\renewcommand{\thefootnote}{\fnsymbol{footnote}}
\footnotetext[2]{Laboratoire d'Informatique Fondamentale de Marseille, Centre National de la Recherche Scientific (LIF - CNRS - UMR 6166)}%
\footnotetext[3]{Institute of Informatics and Telecommunications, NCSR ``Demokritos'', Athens, Greece.}%
\footnotetext[4]{Department of Informatics and Telecommunications, National and Kapodistrian University of Athens, Athens, Greece.}
\renewcommand{\thefootnote}{\arabic{footnote}}
\title{Using Synchronic and Diachronic Relations for Summarizing Multiple Documents Describing Evolving Events}
\date{}

\maketitle

\begin{abstract}
In this paper we present a fresh look at the problem of summarizing evolving events from multiple sources. After a discussion concerning the nature of evolving events we introduce a distinction between \emph{linearly} and \emph{non-linearly} evolving events. We present then a general methodology for the automatic creation of summaries from evolving events. At its heart lie the notions of \emph{Synchronic} and \emph{Diachronic} cross-document Relations (SDRs), whose aim is the identification of similarities and differences between sources, from a synchronical and diachronical perspective. SDRs do not connect documents or textual elements found therein, but structures one might call \emph{messages}. Applying this methodology will yield a set of messages and relations, SDRs, connecting them, that is a graph which we call \emph{grid}. We will show how such a grid can be considered as the starting point of a Natural Language Generation System. The methodology is evaluated in two case-studies, one for linearly evolving events (descriptions of football matches) and another one for non-linearly evolving events (terrorist incidents involving hostages). In both cases we evaluate the results produced by our computational systems.
\end{abstract}

\section{Introduction}
Exchange of information is vital for the survival of human beings. It has taken many forms throughout the history of mankind ranging from gossiping \cite{Pinker:mind} to the publication of news via highly sophisticated media. Internet provides us with new perspectives, making the exchange of information not only easier than ever, but also virtually unrestricted.

Yet, there is a price to be paid to this richness of means, as it is difficult to assimilate this plethora of information in a small amount of time. Suppose a person would like to keep track of the evolution of an event via its description available over the Internet. There is such a vast body of data (news) relating to the event that it is practically impossible to read all of them and decide which are really of interest. A simple visit at, let's say, Google News\footnote{\url{http://www.google.com/news}} will show that for certain events the number of hits, \ie related stories, amounts to the thousands. Hence it is simply impossible to scan through all these documents, compare them for similarities and differences, while reading through in order to follow the evolution of the event.

Yet, there might be an answer to this problem: automatically produced (parametrizable) text summaries. This is precisely the issue we will be concerned with in this paper. We will focus on \emph{Evolving Summarization}; or, to be more precise, the automatic summarization of events evolving throughout time.

\bigskip\noindent
While there has been pioneering work on automatic text summarization  more than 30 years ago, (\citeNP{Luhn58} and \citeNP{Edmundson69}), the field came to a virtual halt until the nineties. It is only then that a revival has taken place (see, for example, \citeNP{Mani&Maybury99,Mani01,Afantenos&al.05:MedicalSurvey} for various overviews). Those early works were mostly concerned with the creation of text summaries from a single source. \mds(MDS) wouldn't be actively pursued until after the mid-1990's --- since when it is a quite active area of research.

Despite its youth, a consensus has emerged within the research community concerning the way to proceed in order to solve the problem. What seems to be at the core of \mds is the identification of similarities and differences between related documents (\citeNP{Mani&Bloedorn99,Mani01}; see also \citeNP{Endres-Niggemeyer98} and \citeNP{Afantenos&al.05:MedicalSurvey}). This  is generally translated as the identification of informationally equivalent passages in the texts. In order to achieve this goal/state, researchers use various methods ranging from statistical \cite{Goldstein&al00}, to syntactic \cite{Barzilay&al99} or semantic approaches \cite{Radev&McKeown98}.

Despite this consensus, most researchers do not know precisely what they mean when they refer to these similarities or differences. What we propose here is that, at least for the problem at hand, \ie of the summarization of evolving events, we should view the identification of the similarities and differences on two axes: the \emph{synchronic} and \emph{diachronic} axis. In the former case we are mostly concerned with the relative agreement of the various sources, within a \emph{given time frame}, whilst in the latter case we are concerned with the actual evolution of an event, as it is being described by a single source.

Hence, in order to capture these similarities and differences we propose to use, what we call, the \emph{Synchronic} and \emph{Diachronic Relations} (henceforth \sdrsshort) across the documents. The seeds of our \sdrsshort lie of course in \citeANP{Mann&Thompson88}'s \citeyear{Mann&Thompson87,Mann&Thompson88} \emph{\rst}(RST). While \rst will be more thoroughly discussed in section~\ref{sec:RelatedWork}, let us simply mention here that it was initially developed in the context of ``computational text generation'',\footnote {Also referred to as \nlg(NLG).} in order to relate a set of small text segments (usually clauses) into a larger, rhetorically motivated whole (text). The relations in charge of gluing the chunks (text segments) are semantic in nature, and they are supposed to capture the authors' (rhetorical) intentions, hence their name.\footnote{In fact, the opinions concerning what \rst relations are supposed to represent, vary considerably. According to one view, they represent the author's intentions; while according to another, they represent the effects they are supposed to have on the readers. The interested reader is strongly advised to take a look at the original papers by \citeANP{Mann&Thompson88} (\citeyearNP{Mann&Thompson87,Mann&Thompson88}), or at \citeN{Taboada&Mann:RST1}.}

\sdrs(SDRs) are similar to \rst relations in the sense that they are supposed to capture similarities and differences, \ie the semantic relations, holding between conceptual chunks, of the input (documents), on the synchronic and diachronic axis. The question is, what are the \emph{units of analysis} for the \sdrs? Akin to work in NLG we could call these chunks ``messages''. Indeed, the initial motivation for \sdrs was the belief or hope that the semantic information they carry could be exploited later on by a generator for the final creation of the summary.

\bigskip\noindent
In the following sections, we will try to clarify what messages and \sdrs are, as well as provide some formal definitions. However, before doing so, we will present in section~\ref{sec:KindsOfEvolution} a discussion concerning the nature of events, as well as a distinction between \emph{linearly} and \emph{non-linearly} evolving events. Section~\ref{sec:Overview} provides a general overview of our approach, while section~\ref{sec:Relations} contains an in-depth discussion of the \sdrslong. In sections~\ref{sec:LinearEvolution} and~\ref{sec:NonLinearEvolution} we present two concrete examples of systems we have built for the creation of Evolving Summaries in a \emph{linearly} and \emph{non-linearly} evolving topic. Section~\ref{sec:NLG} provides a discussion concerning the relationship/relevance of our approach with a \nlglong system, effectively showing how the computational extraction of the messages and \sdrs can be considered as the first stage, out of three, of a typically pipelined \nlg system. Section~\ref{sec:RelatedWork} presents related work, focusing on the link between our theory and \rstlong. In section~\ref{sec:Conclusions} we conclude, by presenting some thoughts concerning future research.

\section{Some Definitions}\label{sec:KindsOfEvolution}
This work is about the summarization of events that evolve through time. A natural question that can arise at this point is \emph{what is}
an event, and \emph{how} do events evolve? Additionally, for a particular
event, do all the sources follow its evolution or does each one have a
different rate for emitting their reports, possibly aggregating several
activities of the event into one report? Does this evolution of the events
affect the summarization process?

\bigskip\noindent
Let us first begin by answering the question of ``what is an event?'' In the
\tdt(TDT) research, an event is described as ``something that happens at some
specific time and place'' (\citeNP{Papka99:PhD}, p~3; see also
\citeNP{Allan&al98:TDTFinRep}). The inherent notion of time is what
distinguishes the event from the more general term \emph{topic}. For example,
the general class of terrorist incidents which include hostages is regarded
as a topic, while a particular instance of this class, such as the one
concerning the two Italian women that were kept as hostages by an Iraqi group
in 2004, is regarded as an event. In general then, we can say that a topic is a \emph{class} of events while an event is an \emph{instance} of a particular topic.

An argument that has been raised in the \tdt research is that although the definition of an event as ``something that happens at some specific time and place'' serves us well in most occasions, such a definition does have some problems \cite{Allan&al98:online}. As an example, consider the occupation of the Moscow Theater in 2002 by Chechen extremists. Although this occupation spans several days, many would consider it as being a single event, even if it does not strictly happen at some ``specific time''. The consensus that seems to have
been achieved among the researchers in \tdt is that events indeed exhibit
evolution, which might span a considerable amount of time
\cite{Papka99:PhD,Allan&al98:online}. \citeN{Cieri.00}, for example, defines an event to be as ``a specific thing that happens at a specific time and place along with all necessary preconditions and unavoidable consequences'', a definition which tries to reflect the evolution of an event.

Another distinction that the researchers in \tdt make is that of the
\emph{activities}. An activity is ``a connected set of actions that have a
common focus or purpose'' \cite[p~3]{Papka99:PhD}. The notion of activities is
best understood through an example. Take for instance the topic of terrorist
incidents that involve hostages. A specific event that belongs to this topic
is composed of a sequence of activities, which could, for example, be the fact
that the terrorists have captured several hostages, the demands that the
terrorists have, the negotiations, the fact that they have freed a hostage,
etc. Casting a more close look on the definition of the activities, we will see
that the activities are further decomposed into a sequence of more simple
\emph{actions}. For example, such actions for the activity of the negotiations
can be the fact that a terrorist threatens to kill a specific hostage unless
certain demands are fulfilled, the possible denial of the negotiation team to
fulfil those demands and the proposition by them of something else, the freeing
of a hostage, etc. In order to capture those actions, we use a structure which
we call \emph{message} --- briefly mentioned in the introduction of this paper.
In our discussion of topics, events and activities we will adopt the
definitions provided by the \tdt research.

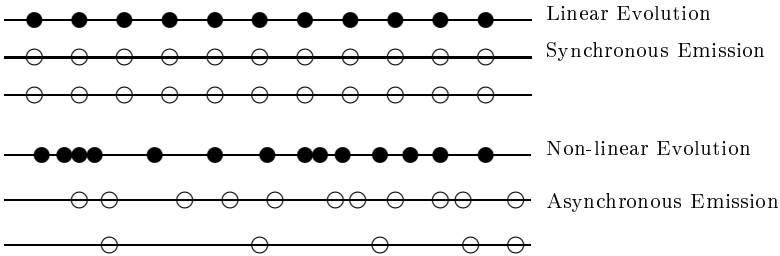
\begin{figure}[hbt]
  \begin{center}
    \setlength{\unitlength}{4144sp}%
\begingroup\makeatletter\ifx\SetFigFont\undefined%
\gdef\SetFigFont#1#2#3#4#5{%
  \reset@font\fontsize{#1}{#2pt}%
  \fontfamily{#3}\fontseries{#4}\fontshape{#5}%
  \selectfont}%
\fi\endgroup%
\begin{picture}(4390,1493)(259,-1644)
{\color[rgb]{0,0,0}\thinlines
\put(496,-1051){\circle*{90}}
}%
{\color[rgb]{0,0,0}\put(631,-1051){\circle*{90}}
}%
{\color[rgb]{0,0,0}\put(721,-1051){\circle*{90}}
}%
{\color[rgb]{0,0,0}\put(811,-1051){\circle*{90}}
}%
{\color[rgb]{0,0,0}\put(1171,-1051){\circle*{90}}
}%
{\color[rgb]{0,0,0}\put(1846,-1051){\circle*{90}}
}%
{\color[rgb]{0,0,0}\put(2071,-1051){\circle*{90}}
}%
{\color[rgb]{0,0,0}\put(2161,-1051){\circle*{90}}
}%
{\color[rgb]{0,0,0}\put(2296,-1051){\circle*{90}}
}%
{\color[rgb]{0,0,0}\put(271,-1051){\line( 1, 0){2250}}
}%
{\color[rgb]{0,0,0}\put(2521,-1051){\line( 1, 0){900}}
}%
{\color[rgb]{0,0,0}\put(1531,-1051){\circle*{90}}
}%
{\color[rgb]{0,0,0}\put(2701,-1051){\circle*{90}}
}%
{\color[rgb]{0,0,0}\put(2881,-1051){\circle*{90}}
}%
{\color[rgb]{0,0,0}\put(3151,-1051){\circle*{90}}
}%
{\color[rgb]{0,0,0}\put(2521,-1051){\circle*{90}}
}%
{\color[rgb]{0,0,0}\put(451,-466){\circle{90}}
}%
{\color[rgb]{0,0,0}\put(721,-466){\circle{90}}
}%
{\color[rgb]{0,0,0}\put(991,-466){\circle{90}}
}%
{\color[rgb]{0,0,0}\put(1261,-466){\circle{90}}
}%
{\color[rgb]{0,0,0}\put(1531,-466){\circle{90}}
}%
{\color[rgb]{0,0,0}\put(1801,-466){\circle{90}}
}%
{\color[rgb]{0,0,0}\put(2071,-466){\circle{90}}
}%
{\color[rgb]{0,0,0}\put(2341,-466){\circle{90}}
}%
{\color[rgb]{0,0,0}\put(2611,-466){\circle{90}}
}%
{\color[rgb]{0,0,0}\put(2881,-466){\circle{90}}
}%
{\color[rgb]{0,0,0}\put(3151,-466){\circle{90}}
}%
{\color[rgb]{0,0,0}\put(3151,-691){\circle{90}}
}%
{\color[rgb]{0,0,0}\put(2881,-691){\circle{90}}
}%
{\color[rgb]{0,0,0}\put(2611,-691){\circle{90}}
}%
{\color[rgb]{0,0,0}\put(2341,-691){\circle{90}}
}%
{\color[rgb]{0,0,0}\put(2071,-691){\circle{90}}
}%
{\color[rgb]{0,0,0}\put(1801,-691){\circle{90}}
}%
{\color[rgb]{0,0,0}\put(1531,-691){\circle{90}}
}%
{\color[rgb]{0,0,0}\put(1261,-691){\circle{90}}
}%
{\color[rgb]{0,0,0}\put(991,-691){\circle{90}}
}%
{\color[rgb]{0,0,0}\put(721,-691){\circle{90}}
}%
{\color[rgb]{0,0,0}\put(451,-691){\circle{90}}
}%
{\color[rgb]{0,0,0}\put(451,-241){\circle*{90}}
}%
{\color[rgb]{0,0,0}\put(721,-241){\circle*{90}}
}%
{\color[rgb]{0,0,0}\put(991,-241){\circle*{90}}
}%
{\color[rgb]{0,0,0}\put(1261,-241){\circle*{90}}
}%
{\color[rgb]{0,0,0}\put(1531,-241){\circle*{90}}
}%
{\color[rgb]{0,0,0}\put(1801,-241){\circle*{90}}
}%
{\color[rgb]{0,0,0}\put(2071,-241){\circle*{90}}
}%
{\color[rgb]{0,0,0}\put(2341,-241){\circle*{90}}
}%
{\color[rgb]{0,0,0}\put(2611,-241){\circle*{90}}
}%
{\color[rgb]{0,0,0}\put(2881,-241){\circle*{90}}
}%
{\color[rgb]{0,0,0}\put(3151,-241){\circle*{90}}
}%
{\color[rgb]{0,0,0}\put(721,-1321){\circle{90}}
}%
{\color[rgb]{0,0,0}\put(901,-1321){\circle{90}}
}%
{\color[rgb]{0,0,0}\put(1351,-1321){\circle{90}}
}%
{\color[rgb]{0,0,0}\put(1891,-1321){\circle{90}}
}%
{\color[rgb]{0,0,0}\put(1621,-1321){\circle{90}}
}%
{\color[rgb]{0,0,0}\put(2251,-1321){\circle{90}}
}%
{\color[rgb]{0,0,0}\put(2386,-1321){\circle{90}}
}%
{\color[rgb]{0,0,0}\put(2611,-1321){\circle{90}}
}%
{\color[rgb]{0,0,0}\put(2881,-1321){\circle{90}}
}%
{\color[rgb]{0,0,0}\put(3016,-1321){\circle{90}}
}%
{\color[rgb]{0,0,0}\put(3331,-1321){\circle{90}}
}%
{\color[rgb]{0,0,0}\put(901,-1591){\circle{90}}
}%
{\color[rgb]{0,0,0}\put(1801,-1591){\circle{90}}
}%
{\color[rgb]{0,0,0}\put(2521,-1591){\circle{90}}
}%
{\color[rgb]{0,0,0}\put(3061,-1591){\circle{90}}
}%
{\color[rgb]{0,0,0}\put(3331,-1591){\circle{90}}
}%
{\color[rgb]{0,0,0}\put(2476,-466){\line( 1, 0){945}}
}%
{\color[rgb]{0,0,0}\put(2476,-691){\line( 1, 0){945}}
}%
{\color[rgb]{0,0,0}\put(2476,-241){\line( 1, 0){945}}
}%
{\color[rgb]{0,0,0}\put(2521,-1321){\line( 1, 0){900}}
}%
{\color[rgb]{0,0,0}\put(2521,-1591){\line( 1, 0){900}}
}%
{\color[rgb]{0,0,0}\put(271,-466){\line( 1, 0){2205}}
}%
{\color[rgb]{0,0,0}\put(271,-691){\line( 1, 0){2205}}
}%
{\color[rgb]{0,0,0}\put(271,-241){\line( 1, 0){2205}}
}%
{\color[rgb]{0,0,0}\put(271,-1321){\line( 1, 0){2250}}
}%
{\color[rgb]{0,0,0}\put(271,-1591){\line( 1, 0){2250}}
}%
\put(3511,-241){\makebox(0,0)[lb]{\smash{{\SetFigFont{8}{9.6}{\familydefault}{\mddefault}{\updefault}{\color[rgb]{0,0,0}Linear Evolution}%
}}}}
\put(3511,-1051){\makebox(0,0)[lb]{\smash{{\SetFigFont{8}{9.6}{\familydefault}{\mddefault}{\updefault}{\color[rgb]{0,0,0}Non-linear Evolution}%
}}}}
\put(3511,-466){\makebox(0,0)[lb]{\smash{{\SetFigFont{8}{9.6}{\rmdefault}{\mddefault}{\updefault}{\color[rgb]{0,0,0}Synchronous Emission}%
}}}}
\put(3511,-1366){\makebox(0,0)[lb]{\smash{{\SetFigFont{8}{9.6}{\rmdefault}{\mddefault}{\updefault}{\color[rgb]{0,0,0}Asynchronous Emission}%
}}}}
\end{picture}%
    \caption{Linear and Non-linear evolution}\label{fig:evolution}
  \end{center}
\end{figure}

\bigskip\noindent
Having thus provided a definition of topics, events and activities, let us now
proceed with our next question of ``how do events evolve through time''.
Concerning this question, we distinguish between two types of evolution:
\emph{linear} and \emph{non-linear}. In linear evolution the major activities
of an event are happening in predictable and possibly constant quanta of time.
In non-linear evolution, in contrast, we cannot distinguish any meaningful
pattern in the order that the major activities of an event are happening. This
distinction is depicted in Figure~\ref{fig:evolution} in which the evolution of
two different events is depicted with the dark solid circles.

At this point we would like to formally describe the notion of linearity. As we
have said, an event is composed of a series of activities. We will denote this
as follows:
\begin{equation*}
  E = \{a_{1}, a_{2}, \ldots, a_{n}\}
\end{equation*}
where each activity $a_{i}$ occurs at a specific point in time, which we will
denote as follows:
\begin{gather*}
  |a_{i}|_{\text{time}} = t_i
\end{gather*}
Such an event $E$ will exhibit linear evolution if
\begin{equation}\label{eq:linear}
  \forall\ k \in \{2, 3, \ldots, n\} \,\, \exists \, m \in \mathds{N}:
  |a_{k}|_{\text{time}} - |a_{k-1}|_{\text{time}} = m\!\cdot\!t
\end{equation}
where $t$ is a constant time unit. On all other cases the event $E$ will
exhibit non-linear evolution. As we have said, linearly evolving events reflect
organized human actions that have a periodicity. Take for instance the event of
a specific football championship. The various matches that compose such an
event\footnote{In this case, the topic is Football Championships, while a
particular event could be the French football championship of 2005-2006. We
consider each match to be an activity, since according to the definitions given
by the \tdt it constitutes ``a connected set of actions that have a common
focus or purpose''.} usually have a constant temporal distance between them.
Nevertheless, it can be the case that a particular match might be canceled due,
for example, to the holidays season, resulting thus in an ``empty slot'' in
place of this match. Equation~(\ref{eq:linear}) captures exactly this
phenomenon. Usually the value of $m$ will be 1, having thus a constant temporal
distance between the activities of an event. Occasionally though, $m$ can take
higher values, \eg 2, making thus the temporal distance between two consecutive
activities twice as big as we would normally expect. In non-linearly evolving
events, on the other hand, the activities of the events do not have to happen
in discrete quanta of time; instead they can follow any conceivable pattern.
Thus any event, whose activities do not follow the pattern captured in
Equation~(\ref{eq:linear}), will exhibit non-linear evolution.

Linearly evolving events have a fair proportion in the world. They can range
from descriptions of various athletic events to quarterly reports that an
organization is publishing. In particular we have examined the descriptions of
football matches (\citeNP{Afantenos&al.04:SETN,Afantenos&al.05:RANLP}; see also
section~\ref{sec:LinearEvolution}). On the other hand, one can argue that most
of the events that we find in the news stories are non-linearly evolving
events. They can vary from political ones, such as various international
political issues, to airplane crashes or terrorist events. As a non-linearly
evolving topic, we have investigated the topic of terrorist incidents which
involve hostages (see section~\ref{sec:NonLinearEvolution}).

\bigskip\noindent
Coming now to the question concerning the rate with which the various sources
emit their reports, we can distinguish between \emph{synchronous} and
\emph{asynchronous} emission of reports. In the case of synchronous emission of
reports, the sources publish almost simultaneously their reports, whilst in the
case of asynchronous emission of reports, each source follows its own agenda in
publishing their reports. This distinction is depicted in
Figure~\ref{fig:evolution} with the white circles. In most of the cases, when
we have an event that evolves linearly we will also have a synchronous emission
of reports, since the various sources can easily adjust to the pattern of the
evolution of an event. This cannot be said for the case of non-linear
evolution, resulting thus in asynchronous emission of reports by the various
sources.

Having formally defined the notions of linearly and non-linearly evolving
events, let us now try to formalize the notion of synchronicity as well. In
order to do so, we will denote the description of the evolution of an event
from a source $S_i$ as
\[S_i = \{r_{i1}, r_{i2}, \ldots r_{in}\}\]
or more compactly as
\[S_i = \{r_{ij}\}_{j = 1}^n\]
where each $r_{ij}$ represents the $j$th report from source $S_i$. Each
$r_{ij}$ is accompanied by its publication time which we will denote as
\[|r_{ij}|_{\texttt{pub\_time}}\]
Now, let us assume that we have two sources $S_k$ and $S_l$ which describe the
same event, \ie
\begin{equation}\label{eq:sources}
\begin{split}
S_k = \{r_{ki}\}_{i=1}^n \\
S_l = \{r_{li}\}_{i=1}^m
\end{split}
\end{equation}
This event will exhibit a synchronous emission of reports if and only if
\begin{gather}
m = n \quad \text{ and, }\label{eq:synchronous1}\\
\forall\ i : |r_{ki}|_{\texttt{pub\_time}} =
|r_{li}|_{\texttt{pub\_time}}\label{eq:synchronous2}
\end{gather}
Equation~(\ref{eq:synchronous1}) implies that the two sources have exactly the
same number of reports, while Equation~(\ref{eq:synchronous2}) implies that all
the corresponding reports are published simultaneously. On the other hand, the
event will exhibit non-linear evolution with asynchronous emission of reports
if and only if
\begin{equation}
\exists\ i : |r_{ki}|_{\texttt{pub\_time}} \neq
|r_{li}|_{\texttt{pub\_time}}\label{eq:asynchronous}
\end{equation}
Equation~(\ref{eq:asynchronous}) implies that at least two of the corresponding
reports of $S_k$ and $S_l$ have a different publication time. Usually of
course, we will have more than two reports that will have a different
publication time. Additionally we would like to note that the $m$ and $n$ of
(\ref{eq:sources}) are not related, \ie they might or might not be
equal.\footnote{In the formal definitions that we have provided for the linear
and non-linear evolution of the events, as well as for the synchronous and
asynchronous emission of reports, we have focused in the case that we have two
sources. The above are easily extended for cases where we have more than two
sources.}

\begin{figure}[htb]
\begin{center}
    \includegraphics[width=0.49\textwidth]{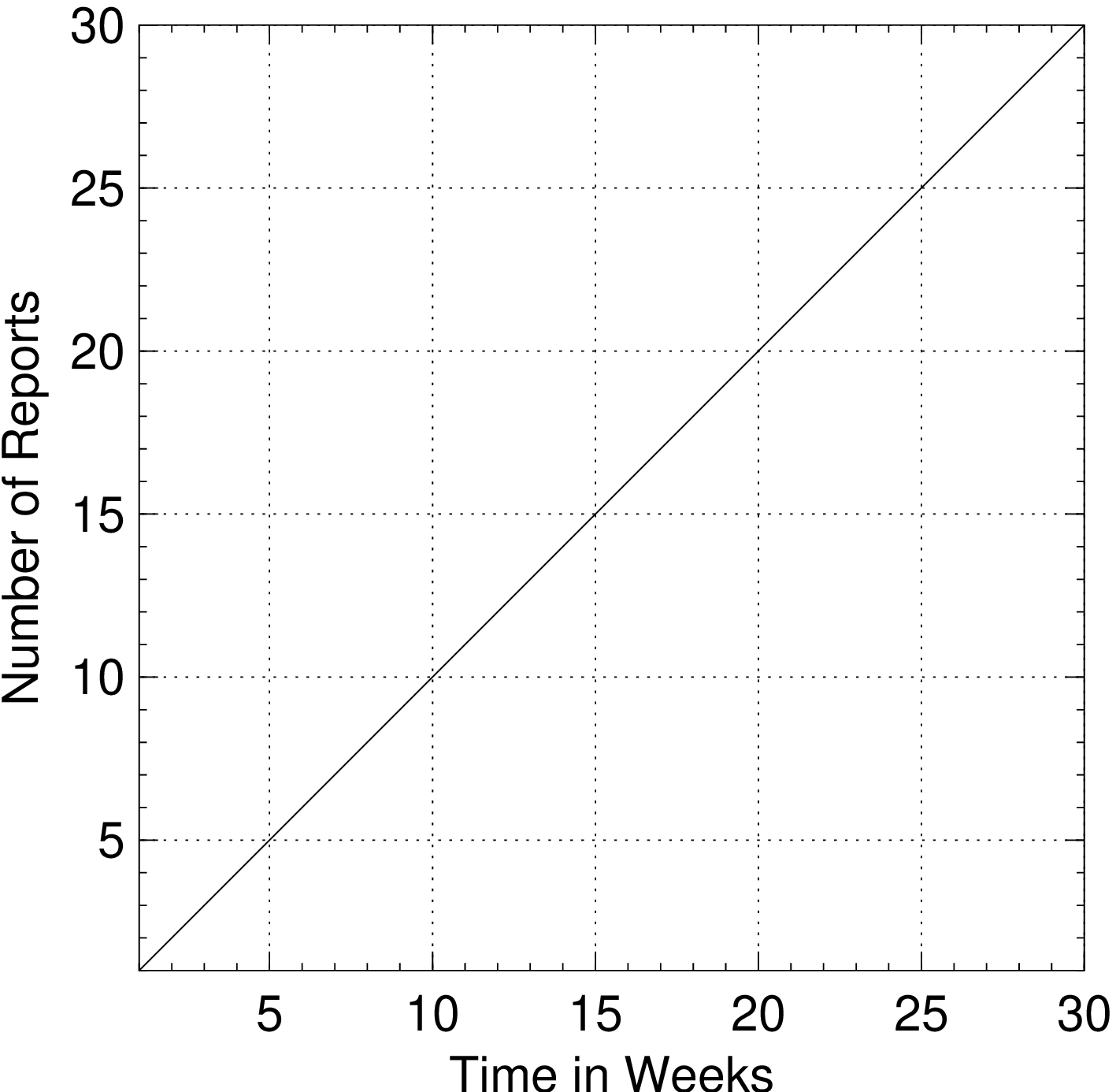}
    \includegraphics[width=0.49\textwidth]{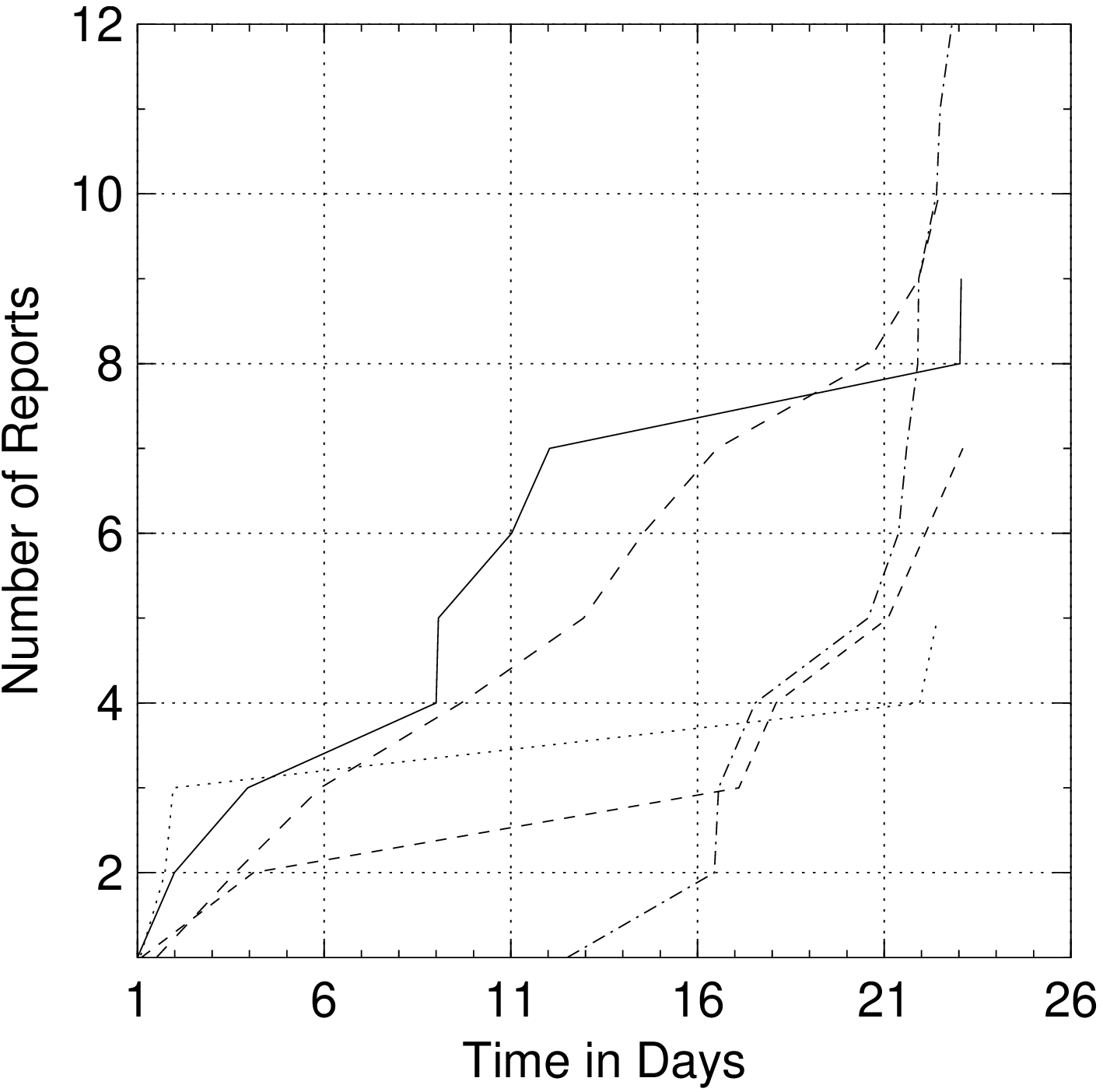}
  \caption{Linear and Non-linear evolution}\label{fig:linearAndNonLinear}
\end{center}
\end{figure}

\bigskip\noindent
In Figure~\ref{fig:linearAndNonLinear} we represent two events which evolve
linearly and non-linearly and for which the sources report synchronously and
asynchronously respectively. The vertical axes in this figure represent the
number of reports per source on a particular event. The horizontal axes
represents the time, in weeks and days respectively, that the documents are
published. The first event concerns descriptions of football matches. In this
particular event we have constant reports weekly from 3 different sources for a
period of 30 weeks. The lines for each source fall on top of each other since
they publish simultaneously. The second event concerns a terrorist group in
Iraq which kept as hostages two Italian women. In the figure we depict 5
sources. The number of reports that each source is making varies from five to
twelve, in a period of about 23 days. As we can see from the figure, most of
the sources begin reporting almost instantaneously, except one which delays its
report for about twelve days. Another source, although it reports almost
immediately, it delays considerably subsequent reports.

\bigskip\noindent
Let us now come to our final question, namely whether the linearity of an event
and the synchronicity of the emission of reports affects our summarization
approach. As it might have been evident thus far, in the case of linear
evolution with synchronous emission of reports, the reports published by the
various sources which describe the evolution of an event, are well aligned in
time. In other words, time in this case proceeds in quanta and in each quantum
each source emits a report. This has the implication that, when the final
summary is created, it is natural that the \nlg component that will create the
text of the summary (see sections~\ref{sec:Overview} and~\ref{sec:NLG}) will
proceed by ``summarizing''\footnote{The word ``summarizing'' here ought to be
interpreted as the \emph{Aggregation} stage in a typical architecture of an
\nlg system. See section~\ref{sec:NLG} for more information on how our approach
is related to \nlg.} each quantum --- \ie the reports that have been published
in this quantum --- separately, exploiting firstly the Synchronic relations for
the identification of the similarities and differences that exist
synchronically for this quantum. At the next step, the \nlg component will
exploit the Diachronic relations for the summarization of the similarities and
differences that exist between the quanta --- \ie the reports published therein
--- showing thus the evolution of the event.

In the case though of non-linear evolution with asynchronous emission of
reports, time does not proceed in quanta, and of course the reports from the
various sources are not aligned in time. Instead, the activities of an event
can follow any conceivable pattern and each source can follow its own agenda on
publishing the reports describing the evolution of an event. This has two
implications. The first is that, when a source is publishing a report, it is
very often the case that it contains the description of many activities that
happened quite back in time, in relation always to the publication time of the
report. This is best viewed in the second part of
Figure~\ref{fig:linearAndNonLinear}. As you can see in this figure, it can be
the case that a particular source might delay the publication of several
activities, effectively thus including the description of various activities
into one report. This means that several of the messages included in such
reports will refer to a point in time which is different from their publication
time. Thus, in order to connect the messages with the \sdrslong the messages
ought to be placed first in their appropriate point in time in which they
refer.\footnote{It could be the case that, even for the linearly evolving
events, some sources might contain in their reports small descriptions of prior
activities from the ones in focus. Although we believe that such a thing is
rare, it is the responsibility of the system to detect such references and
handle appropriately the messages. In the case-study of a linearly evolving
event (section~\ref{sec:LinearEvolution}) we did not identify any such cases.}
The second important implication is that, since there is no meaningful quantum
of time in which the activities happen, then the summarization process should
proceed differently from the one in the case of linear evolution. In other
words, while in the first case the \emph{Aggregation} stage of the \nlg
component (see section~\ref{sec:NLG}) can take into account the quanta of time,
in this case it cannot, since there are no quanta in time in which the reports
are aligned. Instead the \emph{Aggregation} stage of the \nlg component should
proceed differently. Thus we can see that our summarization approach is indeed
affected by the linearity of the topic.

\section{A General Overview}\label{sec:Overview}
As we have said in the introduction of this paper, the aim of this study is to
present a methodology for the automatic creation of summaries from evolving
events. Our methodology is composed of two main phases, the \emph{topic
analysis phase} and the \emph{implementation phase}. The first phase aims at
providing the necessary domain knowledge to the system, which is basically
expressed through an ontology and the specifications of the messages and the
\sdrs. The aim of the second phase is to locate in the text the instances of
the ontology concepts, the messages and the \sdrs, ultimately creating a
structure which we call the \emph{grid}. The creation of the grid constitutes,
in fact, the first stage --- the Document Planning --- out of the three typical
stages of an \nlg system (see section~\ref{sec:NLG} for more details). The
topic analysis phase, as well as the training of the summarization system, is
performed once for every topic, and then the system is able to create summaries
for each new event that is an instance of this topic. In this section we will
elaborate on those two phases, and present the general architecture of a system
for creating summaries from evolving events. During the examination of the
topic analysis phase we will also provide a brief introduction of the notions
of \sdrs, which we more thoroughly present in section~\ref{sec:Relations}. An
in-depth examination on the nature of messages is presented in
section~\ref{sec:Messages}.

\subsection{Topic Analysis Phase}\label{sec:overview:topic}
The topic analysis phase is composed of four steps, which include the creation
of the ontology for the topic, the providing of the specifications for the
messages and the \sdrslong. The final step of this phase, which in fact serves
as a bridge step with the implementation phase, includes the annotation of the
corpora belonging to the topic under examination that have to be collected as a
preliminary step during this phase. The annotated corpora will serve a dual
role: the first is the training of the various \ml algorithms used during the
next phase and the second is for evaluation purposes (see
sections~\ref{sec:LinearEvolution} and~\ref{sec:NonLinearEvolution}). In the
following we will describe in more detail the four steps of this phase. A more
thorough examination of the \sdrslong is presented in
section~\ref{sec:Relations}.

\subsubsection{Ontology}
The first step in the topic analysis phase is the creation of the ontology for
the topic under focus. Ontology building is a field which, during the last
decade, not only has gained tremendous significance for the building of various
natural language processing systems, but also has experienced a rapid
evolution. Despite that evolution, a converged consensus seems to have been
achieved concerning the stages involved in the creation of an ontology
\cite{Pinto&Martins.04,Jones&al.98,Lopez.99}. Those stages include the
\emph{specification}, the \emph{conceptualization}, the \emph{formalization}
and the \emph{implementation} of the ontology. The aim of the first stage
involves the specification of the purpose for which the ontology is built,
effectively thus restricting the various conceptual models used for modeling,
\ie conceptualizing, the domain. The conceptualization stage includes the
enumeration of the terms that represent concepts, as well as their attributes
and relations, with the aim of creating the conceptual description of the
ontology. During the third stage, that conceptual description is transformed
into a formal model, through the use of axioms that restrict the possible
interpretations for the meaning of the formalized concepts, as well as through
the use of relations which organize those concepts; such relations can be, for
example, is-a or part-of relations. The final stage concerns the implementation
of the formalized ontology using a knowledge-representation
language.\footnote{In fact, a fifth stage exists, as well, for the building of
the ontology, namely that of \emph{maintenance}, which involves the periodic
update and correction of the implemented ontology, in terms of adding new
variants of new instances to the concepts that belong to it, as well as its
\emph{enrichment}, \ie the addition of new concepts. At the current state of
our research, this step is not included; nevertheless, see the discussion in
section~\ref{sec:Conclusions} on how this step can, in the future, enhance our
approach.} In the two case-studies of a linearly and non-linearly evolving
topic, which we present in sections~\ref{sec:LinearEvolution}
and~\ref{sec:NonLinearEvolution} respectively, we follow those formal
guidelines for the creation of the ontologies.

\subsubsection{Messages}\label{sec:Messages}
Having provided an ontology for the topic, the next step in our methodology is
the creation of the specifications for the messages, which represent the
actions involved in a topic's events. In order to define what an action is
about, we have to provide a name for the message that represents that action.
Additionally, each action usually involves a certain number of entities. The
second step, thus, is to associate each message with the particular entities
that are involved in the action that this message represents. The entities are
of course taken from the formal definition of the ontology that we provided in
the previous step. Thus, a message is composed of two parts: its \emph{name}
and \emph{a list of arguments} which represent the ontology concepts involved
in the action that the message represents. Each argument can take as value the
instances of a particular ontology concept or concepts, according to the
message definition. Of course, we shouldn't forget that a particular action is
being described by a specific source and it refers to a specific point in time.
Thus the notion of time and source should also be incorporated into the notion
of messages. The source tag of a message is inherited from the source which
published the document that contains the message. If we have a message $m$, we
will denote the source tag of the message as $|m|_\texttt{source}$. Concerning
the time tag, this is divided into two parts: the \emph{publication time} which
denotes the time that the document which contains the message was published,
and the \emph{referring time} which denotes the actual time that the message
refers to. The message's publication time is inherited from the publication
time of the document in which it is contained. The referring time of a message
is, initially, set to the publication time of the message, unless some temporal
expressions are found in the text that alter the time to which the message
refers. The publication and referring time for a message $m$ will be denoted as
$|m|_\texttt{pub\_time}$ and $|m|_\texttt{ref\_time}$ respectively. Thus, a
message can be defined as follows.\footnote{See also
\cite{Afantenos&al.04:SETN,Afantenos&al.05:RANLP,Afantenos&al.05:NLUCS}.}
\begin{center}
\begin{tabularx}{0.7\textwidth}{X}
  $m$\texttt{ = message\_type ( arg$_1$, $\ldots$ , arg$_n$ )}\\
  where \texttt{arg}$_i$ $\in$ Topic Ontology, $i \in \{1,\ldots,n\}$, and:\\
  \\
  $|m|_\texttt{source}$ : the source which contained the message,\\
  $|m|_\texttt{pub\_time}$ : the publication time of the message,\\
  $|m|_\texttt{ref\_time}$ : the referring time of the message.
\end{tabularx}
\end{center}

A simple example might be useful at this point. Take for instance the case of
the hijacking of an airplane by terrorists. In such a case, we are interested
in knowing if the airplane has arrived to its destination, or even to another
place. This action can be captured by a message of type \verb|arrive| whose
arguments can be the entity that arrives (the airplane in our case, or a
vehicle, in general) and the location that it arrives. The specifications of
such a message can be expressed as follows:
\begin{center}
\ttfamily
arrive (what, place)\\
\begin{tabularx}{1.5in}{rX}
  what  :& Vehicle\\
  place :& Location\\
\end{tabularx}
\end{center}
The concepts \verb|Vehicle| and \verb|Location| belong to the ontology of the
topic; the concept \verb|Airplane| is a sub-concept of the \verb|Vehicle|. A
sentence that might instantiate this message is the following:
\begin{quote}
  The Boeing 747 arrived yesterday at the airport of Stanstend.
\end{quote}
For the purposes of this example, we will assume that this sentence was emitted
from source $A$ on 12 February, 2006. The instance of the message is
\begin{center}
\begin{tabularx}{0.8\textwidth}{X}
$m$\texttt{ = arrive ("Boeing 747", "airport of Stanstend")}\\
$|m|_\texttt{source} = A$\\
$|m|_\texttt{pub\_time} = $ \texttt{20060212}\\
$|m|_\texttt{ref\_time} = $ \texttt{20060211}
\end{tabularx}
\end{center}
As we can see, the referring time is normalized to one day before the
publication of the report that contained this message, due to the appearance of
the word ``yesterday'' in the sentence.

The role of the messages' referring time-stamp is to place the message in the
appropriate time-frame, which is extremely useful when we try to determine the
instances of the \sdrslong. Take a look again at the second part of
Figure~\ref{fig:linearAndNonLinear}. As you can see from that figure, there is
a source that delays considerably the publication of its first report on the
event. Inevitably, this first report will try to brief up its readers with the
evolution of the event thus far. This implies that it will mention several
activities of the event that will not refer to the publication time of the
report but much earlier, using, of course, temporal expressions to accomplish
this. The same happens with another source in which we see a delay between the
sixth and seventh report.

At this point, we have to stress that the aim of this step is to provide the
specifications of the messages, which include the provision of the message
types as well as the list of arguments for each message type. This is achieved
by studying the corpus that has been initially collected, taking of course into
consideration the ontology of the topic as well. The actual extraction of the
messages' instances, as well as their referring time, will be performed by the
system which will be built during the next phase. Additionally, we would like
to note that our notion of messages are similar structures (although simpler
ones) to the templates used in the
\muc.\footnote{\url{http://www-nlpir.nist.gov/related_projects/muc/proceedings/muc_7_toc.html}}

\subsubsection{\sdrslong}
Once we have provided the specifications of the messages, the next step in our
methodology is to provide the specifications of the \sdrslong, which will
connect the messages across the documents. Synchronic relations connect
messages from different sources that \emph{refer}\footnote{What we mean by the
use of the word ``refer'' here is that in order to connect two messages with an
SDR we are using their referring time instead of their publication time.} to
the same time frame, while Diachronic relations connect messages from the same
source, but which refer to different time frames. \sdrs are not domain
dependent relations, which implies that they are defined for each topic. In
order to define a relation we have to provide a name for it, which carries
semantic information, and describes the conditions under which this relation
holds, taking into consideration the specifications of the messages. For
example, if we have two different \verb|arrive| messages
\begin{quote}
\verb|            |$m_1$\verb| = arrive (vehicle|$_1$\verb|, location|$_1$\verb|)| \newline%
\verb|            |$m_2$\verb| = arrive (vehicle|$_2$\verb|, location|$_2$\verb|)| %
\end{quote}
and they belong to different sources (\ie $|m_1|_\texttt{source} \neq
|m_2|_\texttt{source}$) but refer to the same time frame (\ie
$|m_1|_\texttt{ref\_time} = |m_2|_\texttt{ref\_time}$) then they will be
connected with the \textsl{Disagreement} Synchronic relation if:
\begin{quote}
\verb|              vehicle|$_1$\verb| = vehicle|$_2$ and\newline%
\verb|              location|$_1 \neq$\verb| location|$_2$%
\end{quote}
On the other hand, if the messages belong to the same source (\ie
$|m_1|_\texttt{source} = |m_2|_\texttt{source}$), but refer to different time
frames (\ie $|m_1|_\texttt{ref\_time} \neq |m_2|_\texttt{ref\_time}$), they
will be connected with the \textsl{Repetition} Diachronic relation if:
\begin{quote}
\verb|              vehicle|$_1$\verb| = vehicle|$_2$ and\newline%
\verb|              location|$_1$\verb| = location|$_2$%
\end{quote}
\sdrslong are more thoroughly examined in section~\ref{sec:Relations}.

\subsubsection{Corpora Annotation}
The fourth and final step in our methodology is the annotation of the corpora,
which ought to have been collected as a preliminary step of this phase. In
fact, this step can be viewed as a ``bridge step'' with the next phase --- the
implementation phase --- since the information that will be annotated during
this step, will be used later in that phase for the training of the various \ml
algorithms, as well as for the evaluation process. In essence, we annotate
three kinds of information during this step. The first is the entities which
represent the ontology concepts. We annotate those entities with the
appropriate ontology (sub)concepts. The next piece of information that we have
to annotate is the messages. This annotation process is in fact split into two
parts. In the first part we have to annotate the textual elements of the input
documents which represent the message types. In the second part we have to
connect those message types with their corresponding arguments. In most of the
cases, as we also mention in sections~\ref{sec:LinearEvolution}
and~\ref{sec:NonLinearEvolution}, we will have an one-to-one mapping from
sentences to message types, which implies that we will annotate the sentences
of the input documents with the appropriate message type. In the second part we
will connect those message types with their arguments, which are in essence the
entities previously annotated. Those entities are usually found in the sentence
under consideration or in the near vicinity of that sentence. Finally we will
have to annotate the \sdrs as well. This is performed by applying the rules
provided in the specification of the Relations (see also
section~\ref{sec:Relations}) to the previously annotated messages.  The
annotation of the entities, messages and \sdrs provides us with a ``gold
corpus'' which will be used for the training of the various \ml algorithms as
well as for the evaluation process.

\subsection{Implementation Phase}\label{sec:overview:impl}
\begin{figure}[htb]
\begin{center}
    \includegraphics[width=\textwidth]{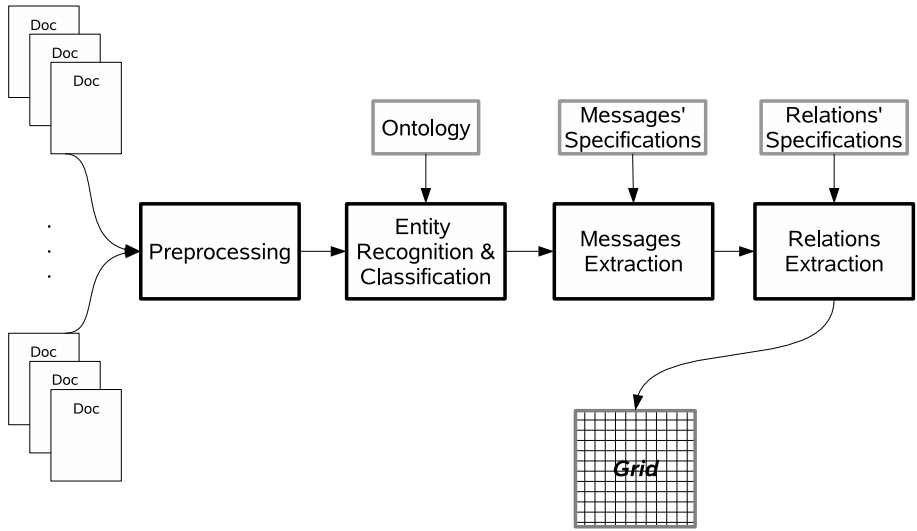}
  \caption{The summarization system.}\label{fig:summarization_core}
\end{center}
\end{figure}

The topic analysis phase is performed once for each topic,\footnote{Although
this is certainly true, in section~\ref{sec:Conclusions} we provide a
discussion on how the system might cope with novel concepts that might arise in
new events that belong to a topic and which have not been included in the
originally created ontology. This discussion is also extended for the case of
messages.} so that the necessary domain knowledge will be provided to the
summarization system which will produce the summaries for each new event that
belongs to this topic. The core of the summarization system is depicted in
Figure~\ref{fig:summarization_core}. As you can see, this system takes as input
a set of documents related to the event that we want to summarize. Those
documents, apart from their text, contain two additional pieces of information:
their source and their publication time. This information will be used for the
determination of the source and publication/referring time of the messages that
are contained in each document. The system is composed of four main stages. In
this section we will briefly mention what the role of each stage is, providing
some clues on the possible computational approaches that can be used. In
sections~\ref{sec:LinearEvolution} and~\ref{sec:NonLinearEvolution} we will
present two concrete computational implementations for a linearly and a
non-linearly evolving topic.

The first stage of the system is a preprocessing that we perform in the input
documents. This preprocessing may vary according to the topic, and it is
actually driven by the needs that have the various \ml algorithms which will be
used in the following stages. In general, this stage is composed of modules
such as a tokenizer, a sentence splitter, a part-of-speech tagger etc. For
example, in the vast majority of cases (as we explain in
sections~\ref{sec:LinearEvolution} and~\ref{sec:NonLinearEvolution}) we had an
one-to-one mapping of sentences to messages. Thus, a sentence splitter is
needed in order to split the document into sentences that will be later
classified into message types. The actual \ml algorithms used will be presented
in sections~\ref{sec:LinearEvolution} and~\ref{sec:NonLinearEvolution}.

\label{discussion:nerc} The next stage of the system is the Entities
Recognition and Classification stage. This stage takes as input the ontology of
the topic, specified during the previous phase, and its aim is to identify the
textual elements in the input documents which denote the various entities, as
well as to classify them in their appropriate (sub)concepts, according to the
ontology. The methods used in order to tackle that problem vary. If, for
example, the entities and their textual realizations are \textit{a priori}
known, then the use of simple gazetteers might suffice. In general though, we
wouldn't normally expect something similar to happen. Thus, a more complex
process, usually including \ml ought to be used for this stage. The identified
entities will later be used for the filling in of the messages' arguments.

\label{discussion:msgs} The third stage is concerned with the extraction of the
messages from the input documents. The aim of this stage is threefold, in fact.
The first thing that should be done is the mapping of the sentences in the
input documents to message types. In the two case studies that we have
performed, and which are more thoroughly described in
sections~\ref{sec:LinearEvolution} and~\ref{sec:NonLinearEvolution}, we came to
the conclusion that in most of the cases, as mentioned earlier, we have an
one-to-one mapping from sentences to message types. In order to perform the
mapping, we are training \ml based classifiers. In
sections~\ref{sec:LinearEvolution} and~\ref{sec:NonLinearEvolution} we will
provide the full details for the two particular topics that we have studied.
The next thing that should be performed during this stage is the filling in of
the messages' arguments; in other words, the connection of the entities
identified in the previous stage with the message types. We should note that,
in contrast with the mapping of the sentences to message types, in this case we
might find several of the messages' arguments occurring in previous or even
following sentences, from the ones under consideration. So, whatever methods
used in this stage, they should take into account not only the sentences
themselves, but their vicinity as well, in order to fill in the messages'
arguments. The final task that should be performed is the identification of the
temporal expressions in the documents that alter the referring time of the
messages. The referring time should be normalized in relation to the
publication time. Note that the publication time and the source tags of the
messages are inherited from the documents which contain the messages.

The final stage in the summarization system is the extraction of the \sdrslong
connecting the messages. This stage takes as input the relations'
specifications and ``interprets'' them into an algorithm which takes as input
the extracted messages, along with their source and publication/referring time
which are attached to the messages. Then this algorithm is applied to the
extracted messages from the previous stage, in order to identify the \sdrs that
connect them. The result of the above stages, as you can see in
Figure~\ref{fig:summarization_core} will be the creation of the structure that
we have called grid.

\begin{figure}[thb]
  \begin{center}
      \includegraphics{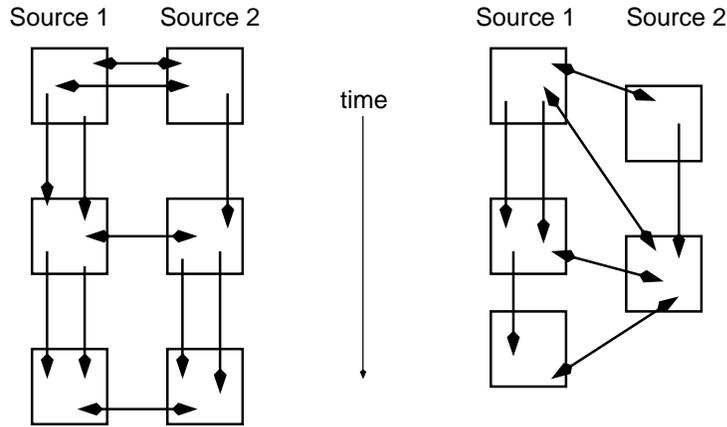}
  \end{center}
  \caption{The grid structure with Synchronic and Diachronic relations for
  linearly and non-linearly evolving events.}
  \label{fig:grid}
\end{figure}

\bigskip\noindent
The \emph{grid} is a structure which virtually provides a level of abstraction
over the textual information of the input documents. In essence, the grid is
composed of the extracted messages, as well as the \sdrslong that connect them.
A graphical representation of two grids, for a linearly evolving event with
synchronous emission of reports and for a non-linearly evolving event with
asynchronous emission of reports respectively, can be seen in
Figure~\ref{fig:grid}. In this figure the squares represent the documents that
the sources emit, while the arrows represent the \sdrslong that connect the
messages which are found inside the documents. In both cases, Synchronic
relations connect messages that belong in the same time-frame,\footnote{A
discussion of what we mean by the ``same time-frame'' can be found in
section~\ref{sec:Relations}. For the moment, suffice it to say that the same
time frame can vary, depending on the topic. In
sections~\ref{sec:LinearEvolution} and~\ref{sec:NonLinearEvolution} we provide
more details for the choices we have made for two different case studies.} but
in different sources, while Diachronic relations connect messages from
different time-frames, but which belong in the same source. Although this is
quite evident for the case of linear evolution, it merits some explanation for
the case of non-linear evolution. As we can see in the second part of
Figure~\ref{fig:grid}, the Synchronic relations can connect messages that
belong in documents from different time-frames. Nevertheless, as we have also
mentioned in section~\ref{sec:overview:topic} in order to connect two messages
with an SDR we take into account their \emph{referring time} instead of their
publication time. In the case of linear evolution it is quite a prevalent
phenomenon that the publication and referring time of the messages will be the
same, making thus the Synchronic relations neatly aligned on the same
time-frame. In the case, though, of non-linear evolution this phenomenon is not
so prevalent, \ie it is often the case that the publication and referring time
of the messages do not coincide.\footnote{If we cast a look again at the second
part of Figure~\ref{fig:linearAndNonLinear} we will see why this is the case.
As we can see there, several sources delay the publication of their reports.
This implies that they can provide information on several of the past
activities of the events, making thus the messages to have different
publication and referring times.} This has the consequence that several of the
Synchronic relations will look as if they connect messages which belong in
different time-frames. Nevertheless, if we do examine the \emph{referring time}
of the messages, we will see that indeed they belong in the same time-frame.

As we have said, the grid provides a level of abstraction over the textual
information contained in the input documents, in the sense that only the
messages and relations are retained in the grid, while all the textual elements
from the input documents are not being included. The creation of the grid
constitutes, in essence, the first stage, the Document Planning, out of the
three total stages in a typical \nlg architecture \cite{Reiter&Dale2000}. We would like to emphasize here the dynamic nature of the grid, concerning on-going events. It could be the case that the system can take as input a set of documents, from various sources, describing the evolution of an event up to a specific point in time. In such cases, the system will build a grid which will reflect the evolution of an event up to this point. Once new documents are given as input to the system, then the grid will be expanded by including the messages extracted from the new documents, as well as the \sdrs that connect those messages with the previous ones or between them. Thus, the grid itself will evolve through time, as new documents are coming as input to the system, and accordingly the generated summary as well. The connection of the grid with the \nlg is more thoroughly discussed in section~\ref{sec:NLG}.

Finally this \nlg system might as well, optionally, take as input
a query from the user, the interpretation of which will create a
\emph{sub-grid} of the original grid. In this case, the sub-grid, instead of
the original grid, will be summarized, \ie will be transformed into a textual
summary. In case that the user enters a query, then a query-based summary will
be created, otherwise a generic one, capturing the whole evolution of the
event, will be created.\footnote{On the distinction between generic and
query-based summaries see \citeN[p~159]{Afantenos&al.05:MedicalSurvey}.}

\section{\sdrslong}\label{sec:Relations}
The quintessential task in the \mdslong research, as we have already mentioned
in the introduction of this paper, is the identification of similarities and
differences between the documents. Usually, when we have the first activity of
an event happening, there will be many sources that will commence describing
that event. It is obvious that the information the various sources have at this
point will vary, leading thus to agreements and contradictions between them. As
the event evolves, we will possibly have a convergence on the opinions, save
maybe for the subjective ones. \emph{We believe that the task of creating a
summary for the evolution of an event entails the description of its evolution,
as well as the designation of the points of confliction or agreement between
the sources, as the event evolves.} In order to capture the evolution of an
event as well as the conflict, agreement or variation between the sources, we
introduce the notion of \emph{\sdrslong}$\!\!$. Synchronic relations try to
identify the degree of agreement, disagreement or variation between the various
sources, at about the same time frame. Diachronic relations, on the other hand,
try to capture the evolution of an event as it is being described by one
source.

According to our viewpoint, \sdrslong ought to be topic-dependent. To put it
differently, we believe that a ``universal'' taxonomy of relations, so to
speak, will not be able to fulfil the intricacies and needs, in terms of
expressive power,\footnote{We are talking about the ``expressive power'' of an
SDR, since \sdrs are ultimately passed over to an \nlg system, in order to be
expressed in a natural language.} for every possible topic. Accordingly, we
believe that \sdrs ought to be defined for each new topic, during what we have
called in section~\ref{sec:Overview} the ``topic analysis'' phase. We would
like though to caution the reader that such a belief does not imply that  a
small pool of relations which are independent of topic, such as for example
\textsl{Agreement, Disagreement} or \textsl{Elaboration}, could not possibly
exist. In the general case though, \sdrs are topic-dependent.

\bigskip\noindent
As we have briefly mentioned in the introduction of this paper, \sdrslong hold
between two different messages. More formally, a relation definition consists
of the following four fields:
\begin{enumerate}
  \item The relation's type (\ie Synchronic or Diachronic).
  \item The relation's name.
  \item The set of pairs of message types that are involved in the relation.
  \item The constraints that the corresponding arguments of each of the pairs
        of message types should have. Those constraints are expressed using the
        notation of first order logic.
\end{enumerate}
The name of the relation carries \emph{semantic} information which, along with
the messages that are connected with the relation, are later being exploited by
the \nlglong component (see section~\ref{sec:NLG}) in order to produce the
final summary. Following the example of subsection~\ref{sec:overview:topic}, we
would formally define the relations \textsl{Disagreement} and
\textsl{Repetition} as shown in Table~\ref{table:rel}.

\begin{table}[bht]
\begin{tabularx}{\textwidth}{|lX|}
  \hline
    \textbf{Relation Name:} & \textsl{DISAGREEMENT}\\
  \hline
    \textbf{Relation Type:} & Synchronic\\
  \hline
    \textbf{Pairs of messages:} & $\{<$\texttt{arrive, arrive}$>\}$\\
  \hline
    \multicolumn{2}{|l|}{\textbf{Constraints on the arguments:}}\\
\end{tabularx}
\begin{tabularx}{\textwidth}{|X|}
  \quad If we have the following two messages:
  \begin{center}
    \texttt{\textbf{arrive} (vehicle$_1$, place$_1$)}\\
    \texttt{\textbf{arrive} (vehicle$_2$, place$_2$)}
  \end{center}
  \quad then we will have a \textsl{Disagreement} Synchronic relation if:
  \begin{center}
    \texttt{(vehicle$_1$ $=$ vehicle$_2$) $\wedge$ (place$_1$ $\neq$ place$_2$)}
  \end{center}\\
  \hline
\end{tabularx}

\bigskip

\begin{tabularx}{\textwidth}{|lX|}
  \hline
    \textbf{Relation Name:} & \textsl{REPETITION}\\
  \hline
    \textbf{Relation Type:} & Diachronic\\
  \hline
    \textbf{Pairs of messages:} & $\{<$\texttt{arrive, arrive}$>\}$\\
  \hline
    \multicolumn{2}{|l|}{\textbf{Constraints on the arguments:}}\\
\end{tabularx}
\begin{tabularx}{\textwidth}{|X|}
  \quad If we have the following two messages:
  \begin{center}
    \texttt{\textbf{arrive} (vehicle$_1$, place$_1$)}\\
    \texttt{\textbf{arrive} (vehicle$_2$, place$_2$)}
  \end{center}
  \quad then we will have a \textsl{Repetition} Diachronic relation if:
  \begin{center}
    \texttt{(vehicle$_1$ $=$ vehicle$_2$) $\wedge$ (place$_1$ $=$ place$_2$)}
  \end{center}\\
  \hline
\end{tabularx}
\caption{Example of formal definitions for two relations.}\label{table:rel}
\end{table}

\bigskip\noindent
The aim of the Synchronic relations is to capture the degree of agreement,
disagreement or variation that the various sources have for the \emph{same
time-frame}. In order thus to define the Synchronic relations, for a particular
topic, the messages that they connect should belong to different sources, but
refer to the same time-frame. A question that naturally arises at this point
is, what do we consider as the same time-frame? In the case of a linearly
evolving event with a synchronous emission of reports, this is an easy
question. Since all the sources emit their reports in constant quanta of time,
\ie at about the same time, we can consider each emission of reports by the
sources, as constituting an appropriate time-frame. This is not though the case
in an event that evolves non-linearly and exhibits asynchronicity in the
emission of the reports. As we have discussed in section~\ref{sec:Overview}, in
such cases, several of the messages will have a reference in time that is
different from the publication time of the document that contains the message.
In such cases we should impose a time window, in relation to the referring time
of the messages, within which all the messages can be considered as candidates
for a connection with a synchronic relation. This time window can vary from
several hours to some days, depending on the topic and the rate with which the
sources emit their reports. In
sections~\ref{sec:LinearEvolution}~and~\ref{sec:NonLinearEvolution}, where we
present two case-studies on a linearly and a non-linearly evolving topics
respectively, we will more thoroughly present the choices that we have made in
relation to the time window.

The aim of Diachronic relations, on the other hand, is to capture the evolution
of an event as it is being described by \emph{one source}. In this sense then
Diachronic relations do not exhibit the same challenges that the Synchronic
ones have, in relation to time. As candidate messages to be connected with a
Diachronic relation we can initially consider all the messages that belong to
the same source but have a different referring time --- but not the same
publication time since that implies that the messages belong in the same
document, something that would make our relations intra-document, instead of
cross-document, as they are intended.

A question that could arise at this point, concerns the chronological distance
that two messages should have in order to be considered as candidates for a
connection with a Diachronic relation. The distance should, definitely, be more
than zero, \ie the messages should not belong in the same time frame. But, how
long could the chronological distance be? It turns out that it all depends on
the topic, and the time that the evolution of the event spans. Essentially, the
chronological distance in which two messages should be considered as candidates
for a connection with a Diachronic relation, depends on the distance in time
that we expect the actions of the entities to affect later actions. If the
effects are expected to have a ``local'' temporal effect, then we should opt
for a small chronological distance, otherwise we should opt for a long one. In
the case-study for the linearly evolving topic
(section~\ref{sec:LinearEvolution}), we chose to have a small temporal
distance, whilst in the non-linearly evolving topic
(section~\ref{sec:NonLinearEvolution}), we chose to have no limit on the
distance.\footnote{Of course, it should be greater than zero, otherwise a zero
distance would make the relation Synchronic, not Diachronic.} The reason for
those decisions will become apparent on the respective sections.

\bigskip\noindent
Until now, in our discussion of the \sdrslong, we have mainly concentrated on
the role that the source and time play, in order for two messages to be
considered as \emph{candidates} for a connection with either a Synchronic or a
Diachronic relation. In order though to establish an actual relation between
two candidate messages, we should further examine the messages, by taking into
account their types and their arguments. In other words, in order to establish
a relation we should provide some rules that take into account the messages'
types as well as the values of their arguments. In most of the cases, we will
have a relation between two messages that have the same message type, but this
is not restrictive. In fact, in the non-linearly evolving topic that we have
examined (section~\ref{sec:NonLinearEvolution}) we have defined several
Diachronic relations that hold between different types of messages.

Once we have defined the names of the relations and their type, Synchronic or
Diachronic, as well as the message pairs for which they hold, then for each
relation we should describe the conditions that the messages should exhibit.
Those conditions take into account the values that the messages' arguments
have. Since the messages' arguments take their values from the topic ontology,
those rules take into account the actual entities involved in the particular
messages. Examples of such rules are provided in
sections~\ref{sec:LinearEvolution} and~\ref{sec:NonLinearEvolution}.

\section{Case Study I: Linear Evolution}\label{sec:LinearEvolution}
This section presents a case study which examines how our approach is applied
to a linearly evolving topic, namely that of the descriptions of football
matches. The reason for choosing this topic is that it is a rather not so
complex one, which makes it quite ideal as a first test bed of our approach.
This is a linearly evolving topic, since football matches occur normally once a
week. Additionally, each match is described by many sources after the match has
terminated, virtually at the same time. Thus we can consider that this topic
exhibits synchronicity on the reports from the various sources. The linearity
of the topic and synchronous emission of reports is depicted in the first part
of Figure~\ref{fig:linearAndNonLinear} (page~\pageref{fig:linearAndNonLinear}),
where we have the description of football matches from three sources for a
period of 30 weeks. The lines from the three sources fall on top of each other
reflecting the linearity and synchronicity of the topic.

\subsection{Topic Analysis}
The aim of the topic analysis phase, as we have thoroughly analyzed in
section~\ref{sec:overview:topic}, is to collect an initial corpus for analysis,
create the ontology for the topic and create the specifications for the
messages and the relations, as well as the annotation of the corpus.

\subsubsection{Corpus Collection}
We manually collected descriptions of football matches, from three sources, for
the period 2002-2003 of the Greek football championship. The sources we used
were a newspaper (\textit{Ta Nea}, \url{http://digital.tanea.gr}), a web portal
(\textit{Flash}, \url{www.flash.gr}) and the site of one football team
(\textit{AEK}, \url{www.aek.gr}). The language used in the documents was Greek.
This championship contained 30 rounds. We focused on the matches of a certain
team, which were described by three sources. So, in total we collected 90
documents containing 64265 words.

\subsubsection{Ontology Creation}
After studying the collected corpus we created the ontology of the topic,
following the formal guidelines in the field of ontology building, a summary of
which we have presented in section~\ref{sec:overview:topic}. The concepts of
the implemented ontology are connected with is-a relations. An excerpt of the
final ontology can be seen in Figure~\ref{fig:ontologyLinear}.

\begin{figure}[htb]
\footnotesize
\begin{verbatim}
          Person                  Temporal Concept        Degree
            Referee                 Minute                Round
            Assistant Referee       Duration              Card
            Linesman                  First Half            Yellow
            Coach                     Second Half           Red
            Player                    Delays              Team
            Spectators                Whole Match
              Viewers
              Organized Fans
\end{verbatim}
\normalsize \caption{An excerpt from the topic ontology for the linearly
evolving topic} \label{fig:ontologyLinear}
\end{figure}

\subsubsection{Messages' Specifications}
Once we have defined the topic ontology, the next stage is the definition of
the messages' specifications. This process includes two things: defining the
message types that exist in the topic, as well as providing their full
specifications. We concentrated in the most important actions, that is on
actions that reflected the evolution of --- for example --- the performance of
a player, or in actions that a user would be interested in knowing. At the end
of this process we concluded on a set of 23 message types
(Table~\ref{table:msgsLinear}). An example of full message specifications is
shown in Figure~\ref{fig:msgSpecsLinear}. As you can see the arguments of the
messages take their values from the topic ontology.

\begin{table}[htb]
  \ttfamily\small
  \begin{tabularx}{\textwidth}{|l|l|l|l|X|}
    \hline
    Absent & Behavior & Block & Card & Goal\_Cancelation\\
    \hline
    Comeback & Final\_Score & Foul & Injured & System\_Selection\\
    \hline
    Performance & Refereeship & Scorer & Change & Satisfaction\\
    \hline
    Superior & Conditions & Penalty & Win & Opportunity\_Lost\\
    \hline
    Expectations & Hope\_For & \multicolumn{2}{|l|}{Successive\_Victories} & \\
    \hline
  \end{tabularx}
  \rmfamily\normalsize
  \caption{Message types for the linearly evolving topic.}\label{table:msgsLinear}
\end{table}

\begin{figure}[htb]
  \centering
  \textbf{performance} (of\_whom, in\_what, time\_span, value)\\
  \begin{tabularx}{2.38in}{Xll}
    of\_whom   &:& Player or Team\\
    in\_what   &:& Action Area\\
    time\_span &:& Minute or Duration\\
    value      &:& Degree
  \end{tabularx}
  \caption{An example of message specifications for the linearly evolving
  topic.}\label{fig:msgSpecsLinear}
\end{figure}

\subsubsection{Relations' Specifications}
We concluded on twelve cross-document relations, six on the synchronic and six
on the diachronic level (Table~\ref{table:relationsLinear}). Since this was a
pilot-study during which we examined mostly the viability of our methodology,
we limited the study of the cross-document relations, in relations that connect
the \emph{same} message types. Furthermore, concerning the Diachronic
relations, we limited our study to relations that have chronological distance
only one, where one corresponds to one week.\footnote{Chronological distance
zero makes the relations synchronic.} Examples of such specifications for the
message type \texttt{performance} are shown in Figure~\ref{fig:relSpecsLinear}.
In the non-linearly evolving topic, examined in the following section, we have
relations that connect different message types, and we impose no limits on the
temporal distance that the messages should have in order to be connected with a
Diachronic relation.

\begin{table}[htb]
\centering
\begin{tabularx}{0.76\textwidth}{lX}
\textsl{\textbf{Diachronic Relations \hspace{0.2in}}} &
\textsl{\textbf{Synchronic Relations}}\\
\hline
-- \small\textsc{Positive Graduation} & -- \small\textsc{Agreement}\\
-- \small\textsc{Negative Graduation} & -- \small\textsc{Near Agreement}\\
-- \small\textsc{Stability}           & -- \small\textsc{Disagreement}\\
-- \small\textsc{Repetition}          & -- \small\textsc{Elaboration}\\
-- \small\textsc{Continuation}        & -- \small\textsc{Generalization}\\
-- \small\textsc{Generalization}      & -- \small\textsc{Preciseness}\\
\end{tabularx}
\caption{\sdrslong in the linearly evolving topic}\label{table:relationsLinear}
\end{table}

\begin{figure}[thb]

\hrulefill

  \footnotesize
  In the following we will assume that we have two messages of type
  \texttt{performance}:
  \begin{quote}
    performance$_1$ (of\_whom$_1$, in\_what$_1$, time\_span$_1$, value$_1$)\\
    performance$_2$ (of\_whom$_2$, in\_what$_2$, time\_span$_2$, value$_2$)
  \end{quote}
The specifications for the relations are the following:

\scriptsize
\smallskip
\begin{tabularx}{\textwidth}{|l||X|X|}
  \hline
    \textbf{Relation Name:} & \textsl{AGREEMENT} & \textsl{DISAGREEMENT}\\
  \hline
    \textbf{Relation Type:} & Synchronic & Synchronic\\
  \hline
    \textbf{Pairs of messages:} & $\{<$\texttt{performance, performance}$>\}$ &  $\{<$\texttt{performance, performance}$>\}$\\
  \hline
    \textbf{Constraints on the} & \texttt{(of\_whom$_1$ $=$ of\_whom$_2$) $\wedge$} & \texttt{(of\_whom$_1$ $=$ of\_whom$_2$) $\wedge$}\\
    \textbf{arguments:} & \texttt{(in\_what$_1$ $=$ in\_what$_2$) $\wedge$} & \texttt{(in\_what$_1$ $=$ in\_what$_2$) $\wedge$}\\
     & \texttt{(time\_span$_1$ $=$ time\_span$_2$) $\wedge$} & \texttt{(time\_span$_1$ $=$ time\_span$_2$) $\wedge$}\\
     & \texttt{(value$_1$ $=$ value$_2$)} & \texttt{(value$_1$ $\neq$ value$_2$)}\\
  \hline
\end{tabularx}

\smallskip
\begin{tabularx}{\textwidth}{|l||X|X|}
  \hline
    \textbf{Relation Name:} & \textsl{POSITIVE GRADUATION} & \textsl{NEGATIVE GRADUATION}\\
  \hline
    \textbf{Relation Type:} & Diachronic & Diachronic\\
  \hline
    \textbf{Pairs of messages:} & $\{<$\texttt{performance, performance}$>\}$ &  $\{<$\texttt{performance, performance}$>\}$\\
  \hline
    \textbf{Constraints on the} & \texttt{(of\_whom$_1$ $=$ of\_whom$_2$) $\wedge$} & \texttt{(of\_whom$_1$ $=$ of\_whom$_2$) $\wedge$}\\
    \textbf{arguments:} & \texttt{(in\_what$_1$ $=$ in\_what$_2$) $\wedge$} & \texttt{(in\_what$_1$ $=$ in\_what$_2$) $\wedge$}\\
     & \texttt{(time\_span$_1$ $=$ time\_span$_2$) $\wedge$} & \texttt{(time\_span$_1$ $=$ time\_span$_2$) $\wedge$}\\
     & \texttt{(value$_1$ $<$ value$_2$)} & \texttt{(value$_1$ $>$ value$_2$)}\\
  \hline
\end{tabularx}

\footnotesize
\bigskip\noindent
Additionally, the messages should satisfy as well the constraints on the source
and referring time in order to be candidates for a Synchronic or Diachronic
Relation. In other words, the messages $m_1$ and $m_2$ will be candidates for a
Synchronic Relation if
\begin{quote}
  $|m_1|_\texttt{source} = |m_2|_\texttt{source}$\\
  $|m_1|_\texttt{ref\_time} = |m_2|_\texttt{ref\_time}$
\end{quote}
and candidates for a Diachronic Relation if
\begin{quote}
  $|m_1|_\texttt{source} = |m_2|_\texttt{source}$\\
  $|m_1|_\texttt{ref\_time} > |m_2|_\texttt{ref\_time}$
\end{quote}

\hrulefill

\normalsize
  \caption{Specifications of \sdrslong for the linearly evolving topic}
  \label{fig:relSpecsLinear}
\end{figure}

\bigskip\noindent
Having provided the topic ontology and the specifications of the messages and
relations, we proceeded with the annotation of the corpora, as explained in
section~\ref{sec:overview:topic}. We would like to add that the total amount of time required for the topic analysis phase was six months for a part-time work of two people.

\subsection{Implementation}
This phase includes the identification in the input documents of the textual
elements that represent ontology concepts, their classification to the
appropriate ontology concept, as well as the computational extraction of the
messages and \sdrslong. At the end of this process, the grid will be created,
which in essence constitutes the Document Planning stage, the first out of
three of a typical \nlg architecture (see section~\ref{sec:NLG}). Casting a
look again in Figure~\ref{fig:summarization_core}, we can see that the
computational extraction of the grid consists of four stages. In the remaining
of this subsection we will discuss those stages.

\subsubsection{Preprocessing}
The preprocessing stage is quite a simple one. It consists of a tokenization
and a sentence splitting components. The information yielded from this stage
will be used in the Entities Recognition and Classification stage and the
messages' extraction stage. We would like to note that in order to perform this
stage, as well as the following two, we used the \textsc{ellogon} platform
\cite{Petasis&al.02}.\footnote{\url{http://www.ellogon.org}}

\subsubsection{Entities Recognition and Classification}
As we discuss on section~\ref{discussion:nerc}, the complexity of the Entities
Recognition and Classification task can vary, depending on the topic. In the
football topic this task was quite straightforward since all the entities
involved in this topic, such as players and teams, were already known. Thus the
use of simple gazetteer lists sufficed for this topic. In the general case
though, this task can prove to be much more complex, as we discuss in
section~\ref{sec:compuNonLinear} on the non-linearly evolving topic.

\subsubsection{Messages Extraction}
This stage consists of three sub-stages. In the first we try to identify the
message types that exist in the input documents, while in the second we try to
fill in the messages' arguments with the instances of ontology concepts which
were identified in the previous stage. The third sub-stage includes the
identification of the temporal expressions that might exist in the text, and
the normalization of the messages referring time, in relation to the
publication time. In this topic however we did not identify any temporal
expressions that would alter the messages referring time, which was set equal
to the messages' publication time. This is natural to expect, since each
document is concerned only with the description of a particular football match.
We can thus consider that this stage consists effectively from two sub-stages.

Concerning the first sub-stage, \ie the identification of the message types, we
approached it as a classification problem. From a study that we carried out, we
concluded that in most of the cases the mapping from sentences to messages was
one-to-one, \ie in most of the cases one sentence corresponded to one message.
Of course, there were cases in which one message was spanning more than one
sentence, or that one sentence was containing more than one message. We managed
to deal with such cases during the arguments' filling sub-stage.

In order to perform our experiments we used a bag-of-words approach according
to which we represented each sentence as a vector from which the stop-words and
the words with low frequencies (four or less) were removed. We performed four
series of experiments. The first two series of experiments used only
\emph{lexical} features, namely the words of the sentences both stemmed and
unstemmed. In the last two series of experiments we enhanced the vectors by
adding to them \emph{semantic} information as well; as semantic features we
used the \emph{NE types} that appear in the sentence. In each of the vectors we
appended the \emph{class} of the sentence, \ie the type of message; in case a
sentence did not correspond to a message we labeled that vector as belonging to
the class \textsl{None}.

In order to perform the classification experiments we used the \textsc{weka}
platform \cite{WEKA.00}. The \ml algorithms that we used were \emph{Na\"{i}ve
Bayes, LogitBoost} and \emph{SMO}. For the last two algorithms, apart from the
default configuration, we performed more experiments concerning several of
their arguments. For all experiments we performed a \emph{ten-fold
cross-validation} with the annotated corpora that we had. Ultimately, the
algorithm that gave the best results was the SMO with the default configuration
for the unstemmed vectors which included information on the NE types. The fact
that the addition of the NE types increases the performance of the classifier,
is only logical to expect since the NE types are used as arguments in the vast
majority of the messages. On the other hand, the fact that by using the
unstemmed words, instead of their stems, increases the performance of the
classifier is counterintuitive. The reason behind this discrepancy is the fact
that the \textsc{skel} stemmer \cite{Petasis&al.03} that we have used, was a
general-purpose one having thus a small coverage for the topic of football
news.

The final sub-stage is the filling in of the messages' arguments. In order to
perform this stage we employed several domain-specific heuristics. Those
heuristics take into account the constraints of the messages, if such
constraints exist. As we noted above, one of the drawbacks of our
classification approach is that there are some cases in which we do not have an
one-to-one mapping from sentences to messages. During this stage of message
extraction we used heuristics to handle many of these cases.

In Table~\ref{table:msgStatsLinear} we show the final performance of the
messages' extraction stage as a whole, when compared against manually annotated
messages on the corpora used. Those measures concern only the message types,
excluding the class \emph{None} messages.

\begin{table}[thb]
  \centering
  \begin{tabular}{ll}
    Precision &: \quad 91.12\%\\
    Recall    &: \quad 67.79\%\\
    F-Measure &: \quad 77.74\%\\
  \end{tabular}
  \caption{Final evaluation of the messages' extraction stage}
  \label{table:msgStatsLinear}
\end{table}

\subsubsection{Relations Extraction}
The final stage towards the creation of the grid is the extraction of the
relations. As is evident from Figure~\ref{fig:relSpecsLinear}, once we have
identified the messages in each document and placed them in the appropriate
position in the grid, then it is fairly straightforward, through their
specifications, to identify the cross-document relations among the messages. In
order to achieve that, we implemented a system written in Java. This system
takes as input the extracted, from the previous stage, messages and it applies
the algorithm, which represents the specifications of the relations, in order
to extract the \sdrs. Ultimately, through this system we manage to represent
the \emph{grid}, which carries an essential role for our summarization
approach. In section~\ref{sec:NLG} we analyze the fact that the creation of the
grid, essentially, consists the Document Planning stage, which is the first out
of three stages of a typical \nlg system architecture \cite{Reiter&Dale2000}.

Concerning the statistics of the extracted relations, these are presented in
Table~\ref{table:relStatsLinear}. As can be seen from that table, the
evaluation results for the relations, when compared with those of the messages,
are somewhat lower. This fact can be attributed to the argument extraction
subsystem, which does not perform as well as the message classification
subsystem.

\begin{table}[htb]
  \centering
  \begin{tabular}{ll}
    Precision &: \quad 89.06\%\\
    Recall    &: \quad 39.18\%\\
    F-Measure &: \quad 54.42\%\\
  \end{tabular}
  \caption{Recall, Precision and F-Measure on the relations}
  \label{table:relStatsLinear}
\end{table}

\bigskip\noindent
In this section we have examined how our methodology for the creation of
summaries from evolving events, presented in section~\ref{sec:Overview}, is
applied to a linearly evolving topic, namely that of the descriptions of
football matches. As we have said in the introduction of this section, this
topic was chosen for its virtue of not being very complex. It was thus an ideal
topic for a first application of our methodology. In the next section we will
move forward and try to apply our methodology into a much more complex topic
which evolves non-linearly.

\section{Case Study II: Non-linear Evolution}\label{sec:NonLinearEvolution}
The topic that we have chosen for our second case study is the terrorist
incidents which involve hostages. The events that belong to this topic do not
exhibit a periodicity concerning their evolution, which means that they evolve
in a non-linear fashion. Additionally, we wouldn't normally expect the sources
to describe synchronously each event; in contrast, each source follows its own
agenda on describing such events. This is best depicted in the second part of
Figure~\ref{fig:linearAndNonLinear} (page~\pageref{fig:linearAndNonLinear}). In
this graph we have the reports for an event which concerns a terrorist group
in Iraq that kept as hostages two Italian women threatening to kill them,
unless their demands were fulfilled. In the figure we depict 5 sources. The
number of reports that each source is making varies from five to twelve, in a
period of about 23 days.

In this section we will once again describe the topic analysis phase, \ie the
details on the collection of the corpus, the creation of the topic ontology,
and the creation of the specifications for the messages and the relations. Then
we will describe the system we implemented for extracting the instances of the
ontology concepts, the messages and the relations, in order to form the grid.

\subsection{Topic Analysis}\label{sec:textAnlNonLinear}
The aim of the topic analysis phase, as we have thoroughly analyzed in
section~\ref{sec:overview:topic} and followed in the previous case-study, is to
collect an initial corpus for analysis, create the ontology for the topic and
create the specifications for the messages and the relations, as well as the
annotation of the corpus.

\subsubsection{Corpus Collection}
The events that fall in the topic of terrorist incidents that involve
hostages are numerous. In our study we decided to concentrate on five such
events. Those events include the hijacking of an airplane from the Afghan
Airlines in February 2000, a Greek bus hijacking from Albanians in July 1999,
the kidnapping of two Italian reporters in Iraq in September 2004, the
kidnapping of a Japanese group in Iraq in April 2004, and finally the hostages
incident in the Moscow theater by a Chechen group in October 2004. In total we
collected and examined 163 articles from 6 sources.\footnote{The sources we
used were the online versions of news broadcasting organizations: the Greek
version of the BBC (\url{http://www.bbc.co.uk/greek/}), the Hellenic
Broadcasting Corporation (\url{http://www.ert.gr}), the Macedonian Press Agency
(\url{http://www.mpa.gr}); a web portal (\url{http://www.in.gr}); and the
online versions of two news papers: \textit{Eleftherotypia}
(\url{http://www.enet.gr}) and \textit{Ta Nea} (\url{http://www.tanea.gr}).}
Table~\ref{table:HostagesDocuments} presents the statistics, concerning the
number of documents and words contained therein, for each event separately.

\begin{table}[htb]
  \centering
  \begin{tabularx}{0.6\textwidth}{|l|r|X|}
    \hline
    Event & Documents & Words \\
    \hline
    Airplane Hijacking & 33 & 7008\\
    Bus Hijacking & 11 & 12416\\
    Italians Kidnaping & 52 & 21200\\
    Japanese Kidnaping & 18 & 10075\\
    Moscow Theater & 49 & 21189\\
    \hline
  \end{tabularx}
  \caption{Number of documents, and words contained therein, for each event.}
  \label{table:HostagesDocuments}
\end{table}

\subsubsection{Ontology Creation}
As in the previous topic examined, we created the ontology following the formal
guidelines that exist in the field of ontology building, a summary of which we
presented in section~\ref{sec:overview:topic}. The concepts of the implemented
ontology are connected with is-a relations. An excerpt of the final ontology
can be seen in Figure~\ref{fig:ontologyNonLinear}.

\begin{figure}[htb]
\footnotesize
\begin{verbatim}
    Person                     Place                  Vehicle
      Offender                   Location of Conduct    Bus
      Hostage                    Country                Plane
      Demonstrators              City                   Car
      Rescue Team              Armament               Media
      Relatives                  Explosive              Newspaper/Press
      Professional               Gas                    Radio
      Governmental Executive     Gun                    Internet
                                 Tank                   TV
\end{verbatim}
\normalsize \caption{An excerpt from the topic ontology for the non-linearly
evolving topic} \label{fig:ontologyNonLinear}
\end{figure}

\subsubsection{Messages' Specifications}
After the creation of the ontology, our metho\-dology requires that we create
the messages' specifications. We would like to remind again that this process
involves two main stages: providing a list with the messages types, and
providing the full specifications for each message. During this process we
focused, as in the previous topic, on the most important messages, \ie the ones
that we believed the final readers of the summary would be mainly interested
in. The messages also had to reflect the evolution of the event. Some of the
messages that we defined, had a very limited frequency in the corpora examined,
thus we deemed them as unimportant, eliminating them from our messages' pool.
At the end of this process we concluded on 48 messages which can be seen in
Table~\ref{table:msgsNonLinear}. Full specifications for two particular
messages can be seen in Figure~\ref{fig:msgSpecsNonLinear}. The first one is
the \verb|negotiate| message, and its semantic translation is that a person is
negotiating with another person about a specific activity. The second message,
\verb|free|, denotes that a person is freeing another person from a specific
location, which can be either the \texttt{Place} or the \texttt{Vehicle}
ontology concepts. Similar specifications were provided for all the messages.

\begin{table}[htb]
  \ttfamily\small
  \begin{tabularx}{\textwidth}{|l|l|l|l|X|}
    \hline
    free & ask\_for & located & assure & take\_on\_responsibility\\
    \hline
    kill & aim\_at & inform & explode & physical\_condition\\
    \hline
    hold & kidnap & organize & be\_afraid & speak\_on\_the\_phone\\
    \hline
    deny & arrive & announce & pay\_ransom & take\_control\_of\\
    \hline
    enter & arrest & transport & escape\_from & give\_deadline\\
     \hline
    help & armed & negotiate & stay\_parked & block\_the\_way\\
    \hline
    meet & leave & threaten & interrogate & hospitalized\\
    \hline
    start & end & work\_for & give\_asylum & head\_towards\\
    \hline
    put & return & hijack & encircle & prevent\_from\\
    \hline
    lead & accept & trade &  & \\
    \hline
  \end{tabularx}
  \rmfamily\normalsize
  \caption{Message types for the linearly evolving topic.}\label{table:msgsNonLinear}
\end{table}

\begin{figure}[htb]
  \centering\small
  \begin{tabularx}{\textwidth}{XX}
  \textbf{negotiate} (who, with\_whom, about) & \textbf{free} (who, whom, from)\\
  who : Person  & who : Person\\
  whom : Person & whom : Person\\
  about : Activity    & from : Place $\vee$ Vehicle\\
  \end{tabularx}
  \caption{An example of message specifications for the non-linearly evolving
  topic.}\label{fig:msgSpecsNonLinear}
\end{figure}

\subsubsection{Relations' Specifications}\label{discussion:relsNonLinear}
The final step during the topic analysis phase is to provide the specifications
for the \sdrslong. As we have explained in section~\ref{sec:Relations},
Synchronic relations hold between messages that have the same referring time.
In the case study examined in the previous section, we did not have any
temporal expressions in the text that would alter the referring time of the
messages in relation to the publication time. In this topic, we do have such
expressions. Thus, Synchronic relations might hold between distant in time
documents, as long as the messages' referring time is the same.

Concerning the Diachronic relations, in the previous topic we examined only
relations that had temporal distance only one, \ie we examined Diachronic
relations that held only between messages found in documents, from the same
source, that had been published consecutively. In this topic we have relaxed
this requirement. This means that messages which have a distant referring time
can be considered as candidates for a connection with a Diachronic relation.
The reason for doing this is that, in contrast with the previous topic, in this
topic we expect the actions of the entities to have an effect which is not
``localized'' in time, but can affect much later actions. This is a direct
consequence of the fact that the events that belong to this topic, have a short
deployment time, usually some days. In the previous topic, the events spanned
several months.

Another difference is that in the previous topic we examined only relations
that hold between the same message types. In this topic we also examine \sdrs
that connect messages with different message types. In the end of this process
we identified 15 \sdrs which can be seen in
Table~\ref{table:relationsNonLinear}. Examples of actual relations'
specifications can be seen in Figure~\ref{fig:relSpecsNonLinear}.

\begin{table}[hbt]
\centering
\begin{tabularx}{0.8\textwidth}{Xl}
\textsl{\textbf{Synchronic Relations}} & \\
\textsl{\textbf{\footnotesize(same message types)}} & \\
-- \small\textsc{Agreement} & \\
-- \small\textsc{Elaboration} & \textsl{\textbf{Diachronic Relations}}\\
-- \small\textsc{Disagreement} & \textsl{\textbf{\footnotesize(different message types)}}\\
-- \small\textsc{Specification} & -- \small\textsc{Cause}\\
 & -- \small\textsc{Fulfillment}\\
\textsl{\textbf{Diachronic Relations}} & -- \small\textsc{Justification}\\
\textsl{\textbf{\footnotesize(same message types)}} & -- \small\textsc{Contribution}\\
-- \small\textsc{Repetition} & -- \small\textsc{Confirmation}\\
-- \small\textsc{Change of Perspective} & -- \small\textsc{Motivation}\\
-- \small\textsc{Continuation} & \\
-- \small\textsc{Improvement} & \\
-- \small\textsc{Degradation} & \\
\end{tabularx}
\caption{\sdrslong in the non-linearly evolving
topic}\label{table:relationsNonLinear}
\end{table}

\begin{figure}[ht]

\hrulefill

\small%
In the following we will assume that we have the following messages
\begin{quote}
  \ttfamily
  negotiate (who$_{a}$, with\_whom$_{a}$, about$_a$)\\
  free (who$_{b}$, whom$_{b}$, from$_b$)\\
  free (who$_{c}$, whom$_{c}$, from$_c$)
\end{quote}
The specifications for the relations are the following:

\smallskip
\begin{tabularx}{\textwidth}{|l||X|X|}
  \hline
    \textbf{Relation Name:} & \textsl{AGREEMENT} & \textsl{POSITIVE EVOLUTION}\\
  \hline
    \textbf{Relation Type:} & Synchronic & Diachronic\\
  \hline
    \textbf{Pairs of messages:} & $\{<$\texttt{free, free}$>\}$ &  $\{<$\texttt{negotiate, free}$>\}$\\
  \hline
    \textbf{Constraints on the} & \texttt{(who$_{b}$ = who$_{c}$) $\wedge$} & \texttt{(who$_{a}$ $=$ who$_{b}$) $\wedge$}\\
    \textbf{arguments:} & \texttt{(whom$_{b}$ = whom$_{c}$) $\wedge$} & \texttt{(about$_{a}$ $=$ free)}\\
     & \texttt{(from$_b$ = from$_c$) $\wedge$} & \\
  \hline
\end{tabularx}

\bigskip\noindent
Additionally, the messages should satisfy as well the constraints on the source
and referring time in order to be candidates for a Synchronic or Diachronic
Relation. In other words, the messages $m_1$ and $m_2$ will be candidates for a
Synchronic Relation if
\begin{quote}
  $|m_1|_\texttt{source} = |m_2|_\texttt{source}$\\
  $|m_1|_\texttt{ref\_time} = |m_2|_\texttt{ref\_time}$
\end{quote}
and candidates for a Diachronic Relation if
\begin{quote}
  $|m_1|_\texttt{source} = |m_2|_\texttt{source}$\\
  $|m_1|_\texttt{ref\_time} > |m_2|_\texttt{ref\_time}$
\end{quote}

\hrulefill

\normalsize \caption{Specifications of \sdrslong for the non-linearly evolving
topic} \label{fig:relSpecsNonLinear}
\end{figure}

\bigskip\noindent
Once the topic ontology, as well as the specifications of the messages and the
relations had been provided, then we proceeded with the final step of the topic
analysis phase of our methodology, which is annotation of the corpora, as was
explained in section~\ref{sec:overview:topic}.  We would like to add that the total amount of time required for the topic analysis phase was six months for a part-time work of two people.

\subsection{Implementation}\label{sec:compuNonLinear}
Having performed the topic analysis phase, the next phase involves the
computational extraction of the messages and relations that will constitute the
grid, forming thus the Document Planning stage, the first out of three, of a
typical \nlg architecture. As in the previous topic, our implementation is
according to the same general architecture presented in
section~\ref{sec:Overview} (see also Figure~\ref{fig:summarization_core}). The
details though of the implementation differ, due to the complexities that this
topic exhibits. These complexities will become apparent in the rest of this
section.

\subsubsection{Preprocessing}
The preprocessing stage, as in the previous case study, is a quite
straightforward process. It also involves a tokenization and a sentence
splitting component, but in this case study it involves as well a
part-of-speech tagger. The information yielded from this stage will be used in
the entities recognition and classification as well as in the messages'
extraction stages, during the creation of the vectors. We would like to note
again that for this stage, as well as for the next two, the \textsc{ellogon}
platform \cite{Petasis&al.02} was used.

\subsubsection{Entities Recognition and Classification}\label{sec:entitiesExtractionNonLinear}
In the present case-study we do not have just named entities that we would like
to identify in the text and categorize in their respective ontology
concepts, but also general entities, which may or may not be named entities. In
other words, during this stage, we are trying to identify the various textual
elements in the input documents that represent an ontology concept, and
classify each such textual element with the appropriate ontology concept. Take
for instance the word \emph{passengers}. This word, depending on the context,
could be an instance of the sub-concept \texttt{Hostages} of the concept
\texttt{Persons} of the ontology  or it might be an instance of the sub-concept
\texttt{Offenders} of the same ontology concept (see again
Figure~\ref{fig:ontologyNonLinear} for the ontology). It all depends on the
context of the sentence that this word appears in. For example, in the
sentence:
\begin{quote}\small
  The airplane was hijacked and its 159 passengers were kept as hostages.
\end{quote}
the word \emph{passengers} ought to be classified as an instance of the
\texttt{Hostages} ontology concept. In contrast, in the following sentence:
\begin{quote}\small
  Three of the airplane's passengers hijacked the airplane.
\end{quote}
the same word, \emph{passengers}, ought to be classified as an instance of the
\texttt{Offenders} ontology concept. It could be the case that under some
circumstances the word \emph{passengers} did not represent an instance of any
ontology concept at all, for the specific topic, since this word did not
participate in any instance of the messages. This is due to the fact that we
have annotated only the instances of the ontology concepts that participate in
the messages' arguments. In fact, after studying the annotated corpora, we
realized that in many occasions textual elements that instantiated an ontology
concept in one context did not instantiate any ontology concept in
another context. Thus the task of the identification and classification of the
instances of the ontology's concepts, in this case-study is much more complex
than the previous one. In order to solve this problem, gazetteer lists are not
enough for the present case study; more sophisticated methods ought to be used.

For this purpose we used \ml based techniques. We opted in using a
\emph{cascade of classifiers}. More specifically, this cascade of classifiers
consists of three levels. At the first level we used a binary classifier which
determines whether a textual element in the input text is an instance of an
ontology concept or not. At the second level, the classifier takes the
instances of the ontology concepts of the previous level and classifies them
under the top-level ontology concepts (such as \texttt{Person} or
\texttt{Vehicle}). Finally at the third level we had a specific classifier for
each top-level ontology concept, which classifies the instances in their
appropriate sub-concepts; for example, in the \texttt{Person} ontology concept
the specialized classifier classifies the instances into \texttt{Offender,
Hostage}, etc. For all the levels of this cascade of classifiers we used the
\textsc{weka} platform. More specifically we used three classifiers:
\emph{Na\"{i}ve Bayes, LogitBoost} and \emph{SMO}, varying the input parameters
of each classifier. We will analyze each level of the cascade separately.

After studying the annotated corpora, we saw that the textual elements that
represent instances of ontology concepts could consist from one to several
words. Additionally, it might also be the case that a textual element that
represents an instance in one context does not represent an instance in another
context. In order to identify which textual elements represent instances of
ontology concepts, we created a series of experiments which took under
consideration the candidate words and their context. We experimented using from
one up to five tokens of the context, \ie before and after the candidate
textual elements. The information we used were token types,\footnote{The types
of the tokens denote whether a particular token was an uppercase or lowercase
word, a number, a date, a punctuation mark, etc.} part-of-speech types, as well
as their combination. After performing a tenfold cross-validation using the
annotated corpora, we found that the classifier which yielded the best results
was LogitBoost with 150 boost iterations,using only the token types and a
context window of four tokens.

The next level in the cascade of classifiers is the one that takes as input the
instances of ontology concepts found from the binary classifier, and determines
their top-level ontology concept (\eg \texttt{Person, Place, Vehicle}). The
features that this classifier used for its vectors, during the training phase,
were the context of the words, as well as the words themselves. More
specifically we created a series of experiments which took into consideration
one to up to five tokens before and after the textual elements, as well as the
tokens which comprised the textual element. The features that we used were the
token types, the part-of-speech types, and their combination. The classifier
that yielded the best results, after performing a tenfold cross-validation, was
LogitBoost with 100 boost iterations with a context of size one, and using as
features the token types and part-of-speech types for each token.

The final level of the cascade of classifiers consists of a specialized
classifier for each top-level ontology concept, which determines the
sub-concepts in which the instances, classified at the previous level, belong.
In this series of experiments we took as input only the nouns that were
contained in each textual element, discarding all the other tokens. The
combined results from the cascade of classifiers, after performing a tenfold
cross-validation, are shown in Table~\ref{table:Cascade}. The last column in
that table, represents the classifier used in the third level of the cascade.
The parameter \emph{I} in the LogitBoost classifier represents the boost
cycles. For conciseness we present only the evaluation results for each
top-level ontology concept. The fact that the \texttt{Person, Place} and
\texttt{Activity} concepts scored better, in comparison to the \texttt{Media}
and \texttt{Vehicle} concepts, can be attributed to the fact that we did not
have many instances for the last two categories to train the classifier.

\begin{table}[htb]
  \centering
  \begin{tabularx}{0.87\textwidth}{|r|l|l|l|X|}
    \hline
    Class & Precision & Recall & F-Measure & Classifier\\
    \hline
    Person & 75.63\% & 83.41\% & 79.33\% & SMO\\
    Place & 64.45\% & 73.03\% & 68.48\% & LogitBoost (I=700)\\
    Activity & 76.86\% & 71.80\% & 74.25\% & LogitBoost (I=150)\\
    Vehicle & 55.00\% & 45.69\% & 49.92\% & Na\"{i}ve Bayes\\
    Media & 63.71\% & 43.66\% & 51.82\% & LogitBoost (I=150)\\
    \hline
  \end{tabularx}
  \caption{The combined results of the cascade of classifiers}
  \label{table:Cascade}
\end{table}

Finally, we would like to note that apart from the above five concepts, the
ontology contained three more concepts, which had a very few instances, making
it inappropriate to include those concepts into our \ml experiments. The reason
for this is that if we included those concepts in our \ml experiments we would
have the phenomenon of skewed class distributions. Instead we opted in using
heuristics for those categories, during which we examined the context of
several candidate words. The results are shown in
Table~\ref{table:heuristicsOntology}.

\begin{table}[htb]
  \centering
  \begin{tabularx}{0.75\textwidth}{|X|l|l|l|}
    \hline
    Ontology Concept & Precision & Recall & F-Measure\\
    \hline
    Public Institution & 88.11\% & 91.75\% & 89.89\%\\
    Physical Condition & 94.73\% & 92.30\% & 93.50\%\\
    Armament & 98.11\% & 100\% & 99.04\%\\
    \hline
  \end{tabularx}
  \caption{Evaluation for the last three ontology concepts}
  \label{table:heuristicsOntology}
\end{table}

\subsubsection{Messages Extraction}
This stage consists of three sub-stages. At the first one we try to identify
the message types that exist in the input documents, while at the second we try
to fill in the messages' arguments with the instances of the ontology concepts
identified in the previous stage. The third sub-stage includes the
identification of the temporal expressions that might exist in the text, and
the normalization of the messages' referring time, in relation to the
document's publication time.

Concerning the first sub-stage, after studying the corpora we realized that we
had an one-to-one mapping from sentences to message types, exactly as happened
in the previous case-study. We used again \ml techniques to classify sentences
to message types. We commenced our experiments by a bag-of-words approach
using, as in the previous case-study, the combination of \emph{lexical} and
\emph{semantic} features. As lexical features we used the words of the
sentences, both stemmed and unstemmed; as semantic features we used the number
of the instances of each sub-concept that were found inside a sentence. This
resulted in a series of four experiments, in each of which we applied the
Na\"{i}ve Bayes, LogitBoost and SMO algorithms of the \textsc{weka} platform.
Unfortunately, the results were not as satisfactory as in the previous case
study. The algorithm that gave the best results was the SMO using both the
semantic and lexical features (as lexical features it used the unstemmed words
of the sentences). The percentage of the message types that this algorithm
managed to correctly classify were 50.01\%, after performing a ten fold cross
validation on the input vectors. This prompted us to follow a different route
for the message type classification experiments.

The vectors that we created, in this new set of \ml experiments, incorporated
again both \emph{lexical} and \emph{semantic} features. As lexical features we
now used only a fixed number of verbs and nouns occurring in the sentences.
Concerning the semantic features, we used two kinds of information. The first
one was a numerical value representing the number of the top-level ontology
concepts (\texttt{Person, Place}, etc) that were found in the sentences. Thus
the created vectors had eight numerical slots, each one representing one of the
top-level ontology concepts. Concerning the second semantic feature, we used
what we have called \emph{trigger words}, which are several lists of words,
each one ``triggering'' a particular message type. Thus, we allocated six slots
--- the maximum number of trigger words found in a sentence --- each one of
which represented the message type that was triggered, if any. In order to
perform our experiments, we used the \textsc{weka} platform. The algorithms
that we used were again the Na\"{i}ve Bayes, LogitBoost and SMO, varying their
parameters during the series of experiments that we performed. The best results
were achieved with the LogitBoost algorithm, using 400 boost cycles. More
specifically the number of correctly classified message types were 78.22\%,
after performing a ten-fold cross-validation on the input vectors.

The second sub-stage is the filling in of the messages' arguments. In order to
perform this stage we employed several domain-specific heuristics which take
into account the results from the previous stages. It is important to note here
that although we have an one-to-one mapping from sentences to message types, it
does not necessarily mean that the arguments (\ie the extracted instances of
ontology concepts) of the messages will also be in the same sentence. There may
be cases where the arguments are found in neighboring sentences. For that
reason, our heuristics use a window of two sentences, before and after the one
under consideration, in which to search for the arguments of the messages, if
they are not found in the original one. The total evaluation results from the
combination of the two sub-stages of the messages extraction stage are shown in
Table~\ref{table:msgTypesEvalNonLinear}. As in the previous cases, we also used
a tenfold cross-validation process for the evaluation of the \ml algorithms.

\begin{table}[htb]
  \centering
  \begin{tabular}{ll}
    Precision &: \quad 42.96\%\\
    Recall    &: \quad 35.91\%\\
    F-Measure &: \quad 39.12\%\\
  \end{tabular}
  \caption{Evaluation for the messages extraction stage of the non-linearly
  evolving topic.} \label{table:msgTypesEvalNonLinear}
\end{table}

At this point we would like to discuss the results a little bit. Although in
the first sub-stage, the classification of the sentences into message types, we
had 78.22\% of the sentences correctly classified, the results of
Table~\ref{table:msgTypesEvalNonLinear} diverge from that number. As we have
noted earlier, the results of Table~\ref{table:msgTypesEvalNonLinear} contain
the \emph{combined} results from the two sub-stages, \ie the classification of
the sentences into message types as well as the filling in of the messages'
arguments. The main reason for the divergence then, seems to be the fact that
the heuristics used in the second sub-stage did not perform quite as well as
expected. Additionally, we would like to note that a known problem in the area
of Information Extraction (IE) is the fact that although the various modules of
an IE system might perform quite well when used in isolation, their
\emph{combination} in most of the cases yields worst results from the expected
ones. This is a general problem in the area of Information Extraction, which
needs to be dealt with \cite{Grishman.05}.

The last of the three sub-stages, in the messages extraction stage, is the
identification of the temporal expressions found in the sentences which contain
the messages and alter their referring time, as well as the normalization of
those temporal expressions in relation to the publication time of the document
which contains the messages. For this sub-stage we adopted a module which was
developed earlier \cite{Stamatiou.TempEx}. As was mentioned earlier in this
paper, the normalized temporal expressions alter the referring time of the
messages, an information which we use during the extraction of the \sdrslong.

\subsubsection{Relations Extraction}
The final processing stage in our architecture is the extraction of the
\sdrslong. As in the previous case-study, the implementation of this stage is
quite straightforward. All that is needed to be done is the translation of the
relations' specifications into an appropriate algorithm which, once applied to
the extracted messages, will provide the relations that connect the messages,
effectively thus creating the grid.

We implemented this stage in Java, creating a platform that takes as input the
extracted messages, including their arguments and the publication and referring
time. The result of this platform is the extraction of the relations. Those
results are shown on Table~\ref{table:relsEvalNonLinear}. As we can see,
although the F-Measures of the messages extraction stage and the relations
extraction stage are fairly similar, their respective precision and recall
values diverge. This is mostly caused due to the fact that small changes in the
arguments of the messages can yield different relations, decreasing the
precision value. The extracted relations, along with the messages that those
relations connect, compose the grid. In section~\ref{sec:NLG} we will
thoroughly present the relation of the grid with the typical stages of an \nlg
component. In fact, we will show how the creation of the grid essentially
constitutes the Document Planning phase, which is the first out of three of a
typical \nlg architecture \cite{Reiter&Dale2000}. Additionally, in that section
we will provide an example of the transformation of a grid into a textual
summary.

\begin{table}[htb]
  \centering
  \begin{tabular}{ll}
    Precision &: \quad 30.66\%\\
    Recall    &: \quad 49.12\%\\
    F-Measure &: \quad 37.76\%\\
  \end{tabular}
  \caption{Evaluation for the relations extraction stage of the non-linearly
  evolving topic.} \label{table:relsEvalNonLinear}
\end{table}

\section{Generating Natural Language Summaries from the Grid}\label{sec:NLG}
In section~\ref{sec:Overview} we have given an overview of our methodology concerning the automatic creation of summaries from evolving events. The results of its application in two case studies have been presented in sections~\ref{sec:LinearEvolution} and~\ref{sec:NonLinearEvolution}. The core of the methodology addresses the issue of extracting the messages and the \sdrslong from the input documents, creating thus a structure we called \emph{grid}. Throughout this paper we have emphasized the fact that this structure will be passed over to a generator for the creation of the final document, \ie summary. In this section we would like to show more concretely the connection between the grid, \ie a set messages and some \sdrs connecting them, with research in \nlglong. More specifically, we would like to show how a grid might be the first part of the typical three components of a generator.

According to \citeN{Reiter&Dale2000} the architecture of a \nlglong system is
divided into the following three stages.\footnote{Rather than following \possessivecite{Reiter&Dale97} original terminology we will follow the terms they used in \citeN{Reiter&Dale2000}, as they seem to be more widely accepted.}
\begin{enumerate}
  \item \textbf{Document Planning.} This stage is divided into two components:
        \begin{enumerate}
          \item \textit{Content Determination.} The core of this stage is
                the determination of what information should be included in the
                generated text. Essentially, this process involves the choice or creation of a \emph{set of messages}  (\citeANP{Reiter&Dale2000} \citeyearNP{Reiter&Dale97,Reiter&Dale2000}) from the underlying
                sources.
          \item \textit{Content Structuring.} The goal of this stage is the
                ordering of the messages created during the previous step,
                taking into account the communicative goals the to-be generated text is supposed to meet. To this end messages are connected with discourse relations. These latter are generally gleaned from \rstlong.
        \end{enumerate}
  \item \textbf{Micro-Planning.} This element is composed of the following three components:
    \begin{enumerate}
      \item \textit{Lexicalization.} This component involves the selection
      of the words to be used for the expression of the messages and relations.
      \item \textit{Aggregation.} At this stage a decision is made concerning the level and location where a message is supposed to be included: in a same paragraph, a sentence or at the clause level. Furthermore, unnecessary or redundant information is factored out, eliminating thus repetition and making the generated text run more smoothly. This component takes also the relations holding between the messages into account.
      \item \textit{Referring Expressions Generation.} The goal of this stage is to determine the information to be given (noun vs. pronoun) in order to allow the reader to discriminate a given object (the referent) from a set of alternatives (a ``cat'' \textit{vs} ``the cat'' \textit{vs} ``it'').
    \end{enumerate}
    \item \textbf{Surface Generation.} In this, final, stage the actual paragraphs and sentences are created according to the specifications of the previous stage. This is in fact the module where the knowledge about the grammar of the target natural language is encoded.
\end{enumerate}
Having provided a brief summary of the stages involved in a typical generator, we would like now to proceed to show how the creation of the grid might indeed, together with the communicative goal, become the starting point, \ie the \emph{document planning}, of an \nlg system.

According to \citeN{Reiter&Dale97}, the main task of \emph{content determination} resides in the choice of ``the entities, concepts and relations'' from the ``underlying data-sources''. Once this is done, we need to structure them.
\begin{quote}\small
  Having established the entities, concepts, and relations we need make use of,
  we can then define a set of messages which impose structure over these
  elements. \cite[p 61]{Reiter&Dale2000}
\end{quote}
The reader should be aware that the ``relations'' mentioned here are different in nature from the \emph{rhetorical relations} to be established during the \emph{content structuring} stage. This being said, let us now try to translate the above given concepts with the ones of our own research. The ``underlying data-sources'', in our case, are the input documents from the various sources, describing the evolution of an event, which we want to summarize. The ``entities and concepts'' in our case are defined in terms of the topic ontology, to be used later on as arguments of the messages. \possessivecite{Reiter&Dale2000} ``relations'' correspond to our \emph{message types}. The structuring is identical in both cases. Hence we can conclude that the two structures are essentially identical in nature. The creation of the messages, which concludes the \emph{content determination} sub-stage, is performed in our case during the messages extraction step of the implementation phase.

The goal of \emph{content structuring}, the next stage, is to impose some order on the messages selected during the content determination sub-stage, by taking communicative goals into account. This is usually achieved by connecting the messages via so called ``discourse relations''. It should be noted however that:
\begin{quote}
  There is no consensus in the research literature on what specific discourse
  relations should be used in an \nlg system. \cite[p~74]{Reiter&Dale97}
\end{quote}
Nevertheless, according to \citeANP{Reiter&Dale2000} probably the most common set of relations used to establish coherence and achieve rhetorical goals, is the one suggested by the \rstlong of \citeANP{Mann&Thompson87} \citeyear{Mann&Thompson87,Mann&Thompson88}, to which they add that
\begin{quote}
  [\lips] many developers modify this set to cater for idiosyncrasies of their
  particular domain and genre.
\end{quote}
The above are, in fact, fully in line with our decision to connect the created messages not with any conventional, a priori set of discourse relations, but rather with what we have called \emph{Synchronic} and \emph{Diachronic Relations},\footnote{While the \sdrs are by no means a modified set of \rst relations, they were certainly inspired by them. In section~\ref{sec:RelatedWork} we will see their respective similarities and differences, as well as where precisely \sdrs provide some improvements over \rst relations.} the latter providing in essence an ordering of the messages scattered throughout the various input documents. Hence we can say, that this component fulfils the same function as the \emph{content structuring} component of the document planning of the \citeANP{Reiter&Dale2000} model, as it connects messages with \sdrs.

\begin{figure}[bth]
\begin{center}
    \includegraphics{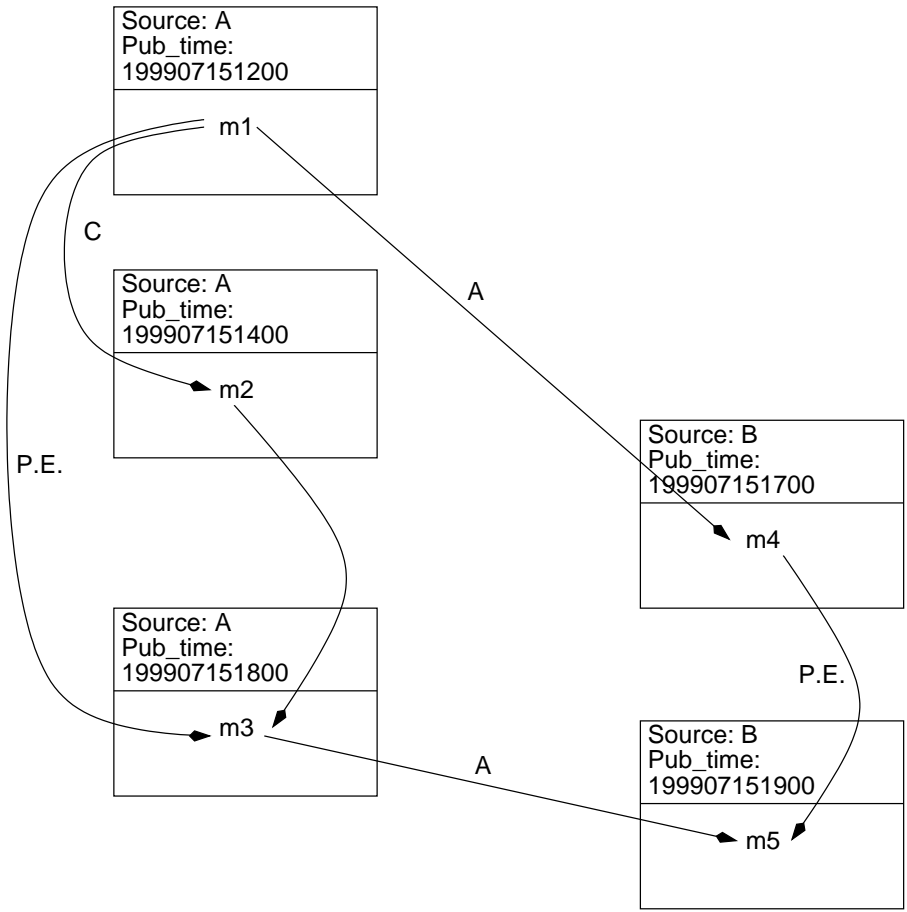}
  \caption{A tiny excerpt of the created grid for the non-linearly evolving
  topic. The messages \textsf{m1--m5} correspond to sentences s$_1$--s$_5$ of
  Table~\ref{table:sentMsgsNonLinear}. \textsf{A}: \textsl{Agreement},
  \textsf{C}: \textsl{Continuation}, \textsf{P.E.}: \textsl{Positive
  Evolution}.}\label{fig:gridNonLinear}
\end{center}
\end{figure}

\begin{table}[htb]
  \begin{tabularx}{\textwidth}{|l|X|}
  \hline
   & Sentence\\
   \hline
  s$_1$ & According to officials, the negotiations between the hijackers and
  the negotiating team have started, and they focus on letting free
  the children from the bus.\\
  s$_2$ & At the time of writing, the negotiations between the hijackers and
  the negotiating team, for the freeing of the children from the bus,
  continue.\\
  s$_3$ & The negotiating team managed to convince the hijackers to let free
  the children from the bus.\\
  s$_4$ & The negotiating team arrived at 12:00 and negotiates with the
  hijackers for the freeing of the children from the bus.\\
  s$_5$ & An hour ago the children were freed from the bus by the
  hijackers.\\
  \hline\hline
   & Message\\
  \hline
  \textsf{m1} & \texttt{negotiate ("negotiating team", "hijackers", "free")}\\
   & |\textsf{m1}|$_{\texttt{source}}$ \texttt{= A}; |\textsf{m1}|$_{\texttt{pub\_time}}$ \texttt{= 199907151200};\\
   & |\textsf{m1}|$_{\texttt{ref\_time}}$ \texttt{= 199907151200}\\
  \textsf{m2} & \texttt{negotiate ("negotiating team", "hijackers", "free")}\\
   & |\textsf{m2}|$_{\texttt{source}}$ \texttt{= A}; |\textsf{m2}|$_{\texttt{pub\_time}}$ \texttt{= 199907151400};\\
   & |\textsf{m2}|$_{\texttt{ref\_time}}$ \texttt{= 199907151400}\\
  \textsf{m3} & \texttt{free ("hijackers", "children", "bus")}\\
   & |\textsf{m1}|$_{\texttt{source}}$ \texttt{= A}; |\textsf{m1}|$_{\texttt{pub\_time}}$ \texttt{= 199907151800};\\
   & |\textsf{m1}|$_{\texttt{ref\_time}}$ \texttt{= 199907151800}\\
  \textsf{m4} & \texttt{negotiate ("negotiating team", "hijackers", "free")}\\
   & |\textsf{m1}|$_{\texttt{source}}$ \texttt{= A}; |\textsf{m1}|$_{\texttt{pub\_time}}$ \texttt{= 199907151700};\\
   & |\textsf{m1}|$_{\texttt{ref\_time}}$ \texttt{= 199907151200}\\
  \textsf{m5} & \texttt{free ("hijackers", "children", "bus")}\\
   & |\textsf{m1}|$_{\texttt{source}}$ \texttt{= A}; |\textsf{m1}|$_{\texttt{pub\_time}}$ \texttt{= 199907151900};\\
   & |\textsf{m1}|$_{\texttt{ref\_time}}$ \texttt{= 199907151800}\\
  \hline
  \end{tabularx}
  \caption{The corresponding sentences and message instances of the
  \textsf{m1--m5} of  Figure~\ref{fig:gridNonLinear}.}
  \label{table:sentMsgsNonLinear}
\end{table}

\noindent%
From the previous analysis we have concluded that the creation of the grid, \ie the identification of messages and their connection with \sdrs, constitutes in essence the first stage of a \nlglong system. Now we would like to show how such a grid can be transformed into a text summary. In Figure~\ref{fig:gridNonLinear} we provide an excerpt from the automatically built grid, of a bus hijacking event, which was thoroughly examined in section~\ref{sec:NonLinearEvolution}.

Each rectangle represents a document annotated with information concerning the source and time of publication of the document. In this small excerpt of the grid we depict one message per source. The messages correspond to the sentences of Table~\ref{table:sentMsgsNonLinear}. They are connected with the Synchronic and Diachronic relations as shown in Figure~\ref{fig:gridNonLinear}. Note that in order to establish a Synchronic relation between two messages reference time is taken into account rather than the time of publication of the messages. The messages for which we have a different reference time, as opposed to their publication time, are \textsf{m4} and \textsf{m5}. This is marked explicitly by the temporal expressions ``at 12:00'' and ``an hour ago'' in sentence s$_4$ and s$_5$. Thus the messages \textsf{m4}, \textsf{m5} and \textsf{m1}, \textsf{m3} are connected respectively via  the Synchronic relation \textsl{Agreement} as: (1) they belong to different sources, (2) they have the same reference time, and (3) their arguments fulfil the constraints presented in Figure~\ref{fig:relSpecsNonLinear}. A similar syllogism applies for the Diachronic relations. Hence, the messages \textsf{m1} and \textsf{m3} are connected via a \textsl{Positive Evolution} Diachronic relation because: (1) they belong to the same source, (2) they have different reference times, and (3) their arguments fulfil the constraints presented in Figure~\ref{fig:relSpecsNonLinear}. Once such a grid is passed to the \nlg component, it may lead to the following output, \ie summary.
\begin{quote}
According to all sources, the negotiations between the hijackers and the
negotiating team, for the freeing of the children, started at 12:00. The
continuous negotiations resulted in a positive outcome at 18:00 when the
hijackers let free the children.
\end{quote}

\section{Related Work}\label{sec:RelatedWork}
In this paper we have presented a methodology which aims at the automatic creation of summaries from evolving events, \ie events which evolve over time and which are being described by more than one source. Of course, we are not the first ones to incorporate directly, or indirectly, the notion of time in our approach of summarization. For example, \citeN{Lehnert.81}, attempts to provide a theory for what she calls \emph{narrative summarization}. Her approach is based on the notion of ``plot units'', which connect \emph{mental states} with various relations, likely to be combined into highly complex patterns. This approach applies for single documents. Bear in mind though that the author does not provide any implementation of her theory. More recently, \citeN{Mani.04} attempts to revive this theory, although, again we lack a concrete implementation validating the approach.

From a different viewpoint, \citeN{Allan&al01} attempt what they call \emph{temporal summarization}. In order to achieve this goal, they start from the results of a \tdtlong system for an event, and order sentences chronologically, regardless of their origin, creating thus a stream of sentences. Then they apply two statistical measures, \emph{usefulness} and \emph{novelty}, to each ordered sentence. The aim being the extraction of sentences whose score is above a given threshold. Unfortunately, the authors do not take into account the document sources, and they do not consider the evolution of the events; instead they try to capture novel information. Actually, what \citeN{Allan&al01} do is to create an extractive summary, whereas we aim at the creation of abstractive summaries.\footnote{Concerning the difference between these two kind of summaries see \citeN[p~160]{Afantenos&al.05:MedicalSurvey}.}

As mentioned already, our work requires some domain knowledge, acquired during the so called ``topic analysis'' phase, which is expressed conjointly via the ontology, and the specification of messages and relations. One such system based on domain knowledge is \textsc{summons} \cite{Radev&McKeown98,Radev99:PhD}. The main domain specific knowledge of this system comes from the specifications of the \muc conferences. \textsc{summons} takes as input several \muc templates and, having applied a series of operators, it tries to create a baseline summary, which is then enhanced by various named entity descriptions collected from the Internet. Of course, one could argue that the operators used by \textsc{summons} resemble our \sdrs. However, this resemblance is only superficial, as our relations are divided into \emph{Synchronic} and \emph{Diachronic} ones, thus reporting similarities and differences in two opposing directions.

Concerning the use of relations, there have been several attempts in the past to try to incorporate them, in one form or another, in summary creation. \citeN{Salton&al97}, for example, try to extract paragraphs from a single document by representing them as vectors and assigning a relation between the vectors if their similarity exceeds a certain threshold. They present then various heuristics for the extraction of the best paragraphs.

Finally, \citeN{Radev00} proposed the \cstlong taking into account 24 domain independent relations existing between various text units across documents. In a later paper \citeN{Zhang&al02} reduce the set to 17 relations and perform some experiments with human judges. These experiments produce various interesting results. For example, human judges annotate only sentences, ignoring completely any other textual unit (phrases, paragraphs, documents) suggested by the theory. Also, the agreement between judges concerning the type of relation holding between two connected sentences is rather small. Nevertheless, \citeN{Zhang&al03} and \citeN{Zhang&Radev.04a} continued to explore these issues by using \ml algorithms to identify cross-document relations. They used the Boosting algorithm and the F-measure for evaluation. The results for six classes of relation, vary from 5.13\% to 43.24\%. However, they do not provide any results for the other 11 relations.\footnote{By contrast, in our work the F-Measure for \emph{all} the relations, is 54.42\% and 37.76\% respectively for the topics of the football matches and the terrorist incidents involving hostages.} Concerning the relations we should note, that, while a general \emph{pool} of cross-document relations might exist, we believe that, in contrast to \citeN{Radev00}, they are domain dependent, as one can choose from this pool the appropriate subset of relations for the domain under consideration, possibly enhancing them with completely domain specific relations to suit one's own needs. Another significant difference from our work, is that we try to create summaries that show not only the evolution of an event, but also the similarities or differences of the sources during the event's evolution.

\bigskip\noindent
Another kind of related work that we would like to discuss here is the \rstlong(RST). Although \rst has not been developed with automatic text summarization in mind, it has been used by \citeANP{Marcu2000} \citeyear{Marcu97,Marcu2000} for the creation of extractive single-document summaries. In this section we will not discuss \citeANP{Marcu2000}'s work since it concerns the creation of summaries from single documents.\footnote{The interested reader should take a look at his works (\eg \citeANP{Marcu2000} \citeyearNP{Marcu97,Marcu2000,Marcu01}). For a comparison of this and other related works you may consider taking a look at \citeN{Mani01} or \citeN{Afantenos&al.05:MedicalSurvey}.} Instead, in the following we will attempt a comparison of our approach with \rst, specifying their respective similarities and differences, as well as the points where our approach presents an innovation with regard to \rst. We would like though to issue a warning to the reader that, even if we claim that our approach extends the \rstlong, we are fully aware of our intellectual debts towards the authors of \rst. The innovations we are claiming here are somehow linked to the specific context of summarizing evolving events. In fact, the decisions we have made have recently found a kind of assent by one of the creators of \rst in a paper entitled ``Rhetorical Structure Theory: Looking Back and Moving Ahead'' \cite{Taboada&Mann:RST1}. What we mean by this is that what the authors provide as innovations to be considered in the future of \rst, have, in a sense, been implemented by us, be it though in the context of the text summarization of evolving events. This being said, let us proceed with a brief description of \rst and the similarities, differences and innovations of our work.

\rstlong has been introduced by \citeANP{Mann&Thompson87} \citeyear{Mann&Thompson87,Mann&Thompson88}. It was originally developed to address the issue of text planning, or text structuring in \nlg,  as well as to provide a more general theory of how coherence in texts is achieved \cite{Taboada&Mann:RST1}. This theory made use of a certain number of \emph{relations}, which carried semantic information. Examples of such relations are \textsl{Contrast, Concession, Condition}, etc. The initially proposed set contained 24 relations \cite{Mann&Thompson88}; today we have 30 relations \cite{Taboada&Mann:RST1}. Each relation holds between two or more segments, \emph{units of analysis}, generally clauses. The units, schemata, are divided into \emph{nuclei} and \emph{satellites}, depending on their relative importance. Only the most prominent part, the nucleus, is obligatory. Relations can hold not only between nuclei or satellites but also between any of them and an entire schema (a unit composed of a nucleus and a satellite), hence, potentially we have a tree.

As mentioned already, \rst was developed, with the goal of \nlglong: ``It was intended for a particular kind of use, to guide computational text generation'' \cite[p~425]{Taboada&Mann:RST1}. In fact, this is also what we had in mind, when we developed our approach. As explained in section~\ref{sec:Messages}, our notion of  messages was inspired by the very same notion used in the domain of \nlg. In addition, our messages are connected with \sdrslong, forming thus what we have called a \emph{grid}, that is a structure to be handed over to the surface generation component of an \nlg system in order to create the final summary.

The point just made is one of similarity between the two approaches. Let us now take a look at a point where we believe to be innovative. As mentioned already, \rst relations hold generally between clauses.\footnote{And, of course, between \emph{spans} of units of analysis.} As \citeN{Taboada&Mann:RST1} write, choosing the clause as the unit of analysis works well in many occasions; but they concede that occasionally, this has some drawbacks. Actually they write (p~430) that:
\begin{quote}\small
  We do not believe that one unit division method will be right for everyone;
  we encourage innovation.
\end{quote}
This is precisely the point where we are innovative. Our units of analysis are not clauses, or any other textual element, rather we have opted for \emph{messages} as the units of analysis, which, as mentioned in section~\ref{sec:Messages}, impose a structure over the entities found in the input text.

While the units of analysis in \rst are divided into nuclei and satellites, our units of analysis --- messages --- do not have such a division. This is indeed a point where \rst and our approach differ radically. In \rst nuclei are supposed to represent more prominent information, compared to satellites. In our own approach this is remedied through the use of a query (see section~\ref{sec:Overview}) from the user. What we mean by this is that we do not \emph{a priory} label the units of analysis in terms of relative importance, instead we let the user do this. In a sense, we determine prominence via a query, which is then thoroughly analyzed by our system, so that it will then be mapped to the messages, and accordingly to the \sdrs that connect them, and describe best the query.

Let us now say a few words concerning the taxonomy of relations. As explained in section~\ref{sec:Relations} we divide our relations into Synchronic and Diachronic relations. In addition we assume that these relations are domain-dependant, in the sense that we have to define \sdrs for each new topic. While a stable set of topic independent \sdrs might exist, we do not make such a claim. By contrast, \rst relations are domain independent. The initial set of \rst relations had a cardinality of 24, with 6 more relations added more recently, which leaves us with 30 \rst relations. This set is considered by many researchers, though not by all, as a fixed set. Yet, this is not what was intended by the developers of \rstlong. As pointed out by \citeN[p~256]{Mann&Thompson88}: ``no single taxonomy seems suitable'', which encourages our decision to have  topic sensitive \sdrs, that is, \sdrs being defined for each new topic, in order to fulfil the needs of each topic. In fact, \citeN[p~438]{Taboada&Mann:RST1} claim that:
\begin{quote}\small
  There may never be a single all-purpose hierarchy of defined relations,
  agreed upon by all. But creating hierarchies that support particular
  technical purposes seems to be an effective research strategy.
\end{quote}
which somehow supports our decisions to introduce topic sensitive \sdrs.

Another point where \rst seems to hold similar views as we do, is the semantics of the relations. In both cases, relations are supposed to carry \emph{semantic} information. In our approach this information will be exploited later on by the generator for the creation of the final summary, whereas in \rst it is supposed to show the coherence of the underlying text and to present the authors' intentions, facilitating thus the automatic generation of text. While the relations carry semantic information in both cases, in \rst they were meant above all to capture the authors' intentions. We do not make such a claim.

A final, probably minor point in which the two approaches differ is the resulting graph. In \rst the relations form a \emph{tree}, whilst in our theory the relations form a \emph{directed acyclic graph}. This graph, whose messages are the vertices and the relations the edges, forms basically what we have called the grid, that is the structure to be handed down to the \nlg component.

\section{Conclusions and Future Work}\label{sec:Conclusions}
In this paper we have presented a novel approach concerning the summarization of multiple documents dealing with evolving events. One point we focused particularly on was the automatic detection of the \sdrslong. As far as we know, this problem has never been studied before. The closest attempt we are aware of is \possessivecite{Allan&al01} work, who create what they call temporal summaries. Nevertheless, as explained in section~\ref{sec:RelatedWork}, this work does not take into account the event's evolution. Additionally, they are in essence agnostic in relation to the source of the documents, since they concatenate all the documents, irrespective of source, into one big document in which they apply their statistical measures.

In order to tackle the problem of summarizing evolving events, we have introduced the notions of messages and \sdrslong(SDRs). Messages impose a structure over the instances of the ontology concepts found in the input texts. They are the units of analysis for which the \sdrs hold. Synchronic relations hold between messages from different sources with identical reference time, whilst Diachronic relations hold between messages from the same source with different reference times. In section~\ref{sec:KindsOfEvolution} we provided definitions for the notions of \emph{topic, event} and \emph{activities}, borrowing from the terminology of \tdtlong research. We also drew a distinction concerning the evolution of the events, dividing them into linear and non-linear events. In addition, we made a distinction concerning the report emission rate of the various sources, dividing them into synchronous and asynchronous emissions. We also provided a formal framework to account for the notions of linearity and synchronicity. Finally, we have shown how these distinctions affect the identification of the \sdrslong.

In section~\ref{sec:Overview} we have presented our methodology behind the implementation of a system that extracts \sdrslong from descriptions of evolving events. This methodology is composed of two phases: the topic analysis phase and the implementation phase, presented in the subsections~\ref{sec:overview:topic} and~\ref{sec:overview:impl} respectively. In sections~\ref{sec:LinearEvolution} and~\ref{sec:NonLinearEvolution} we described two case-studies for a linearly and non-linearly evolving topic, which implement the proposed methodology. While the results are promising in both cases, there is certainly room for improvement for certain components. The tools incorporated for the implementation include the \textsc{weka} platform for the training of the Machine Learning algorithms, as well as the \textsc{ellogon} platform used for the annotation stage of the topic analysis phase and the development of the module used in the extraction of the messages.

In section~\ref{sec:NLG} we have shown how the creation of the grid, \ie the extraction of the messages and their connection via \sdrslong, forms essentially the Document Planning stage, \ie the first out of the three stages of a typical \nlg system \cite{Reiter&Dale2000}. Finally, in section~\ref{sec:RelatedWork} we have presented related works, emphasizing the relationship between \rstlong and our approach. We have shown the respective similarities and differences between the two, highlighting the innovative aspects of our approach. These innovations are in line with what one of the creators of \rst presents as one the points that ought to be considered for the future of \rst in a recent paper entitled ``Rhetorical Structure Theory: Looking Back and Moving Ahead'' \cite{Taboada&Mann:RST1}. Again, we would like though to emphasize that, while certain parts of our approach have been inspired by \rst, the approach as a whole should not be considered as an attempt of improvement of \rst. In a similar vein, our innovations, should not be considered as an extension of \rst. Instead it should merely be viewed as a new kind of methodology to tackle the problem of summarization of evolving events, via \sdrslong.

\bigskip\noindent
As mentioned in section~\ref{sec:Overview} we have presented a general architecture of a system which implements the proposed approach. The implementation of the \nlg subsystem has not been completed yet. The Micro-Planning and Surface Generation stages are still under development. The completion of the \nlg component is an essential aspect of our current work. Even if the results of the entities-, message-, and relation-extraction components --- which are part of the summarization core --- yield quite satisfactory results, we need to \emph{qualitatively} evaluate our summaries. Yet, this will only be possible once the final textual summaries are created, and this requires the completion of the \nlg component.

As shown in the evaluation of the system's components, the results concerning the summarization core are quite promising. Obviously, there is still room for improvement. The component that seems to need most urgent consideration is the arguments filling component. Up to now we are using heuristics which take into account the sentences' message types, returned by the dedicated classifier, as well as the extracted entities, resulting from the various classifiers used (see section~\ref{sec:entitiesExtractionNonLinear}). This method does seem to be brittle, hence additional methods might be needed to tackle this problem. One idea would be to study various \ml methods taking into account previously annotated messages, \ie message types and their arguments. Another module needing improvement is the entity-extraction component, especially the first classifier (the binary classifier) of the cascade of classifiers presented.

Concerning the summarization core, as we have shown in the evaluation of the
several components included in this system, the results are promising. Yet,
there is still room for improvement. The component that seems to need an
immediate consideration is the arguments filling one. Up till now we are using
heuristics which take into consideration the message type of the sentence, as
returned by the dedicated classifier, as well as the extracted entities, which
are in turn the result of the various classifiers used (see
section~\ref{sec:entitiesExtractionNonLinear}). This method does not seem to
perform perfectly, which means that additional methods should be considered in
order to tackle that problem. An idea would be the investigation of various \ml
methods which would take into account previously annotated messages, \ie
message types with their arguments. An additional module that needs improvement
is the entities extraction component, especially the first classifier (the
binary classifier) in the cascade of classifiers that we have presented.

An additional point that we would like to make concerns the nature of messages and the reduction of the human labor involved in the provision of their specifications. As it happens, the message types that we have provided for the two case studies, rely heavily on either verbs or verbalized nouns. This implies that message types could be defined automatically based mostly on statistics on verbs and verbalized nouns. Concerning their arguments, we could take into account the types of the entities that exist in their near vicinities. This is an issue that we are currently working on. Another promising path for future research might be the inclusion of the notion of messages, and possibly the notion of \sdrslong, into the topic ontology.

\relsize{-1}
\section*{Acknowledgments}
The first author was partially supported by a research scholarship from the
Institute of Informatics and Telecommunications of NCSR ``Demokritos'', Athens,
Greece. The authors would like to thank Michael Zock for his invaluable and copious comments on a draft of the paper. Additionally, the authors would like to thank Eleni Kapelou and Irene Doura for their collaboration on the specifications of the messages and relations for the linearly evolving topic, and Konstantina Liontou and Maria Salapata for the collaboration on the specifications of the messages and relations for the non-linearly evolving topic, as well as the annotation they have performed. Finally the authors would like to thank George Stamatiou for the implementation of the temporal expressions module on the non-linearly evolving topic.
\relsize{+1}


\bibliography{JIIS}

\bibliographystyle{achicago}

\end{document}